\documentclass{article}

\usepackage{microtype}
\usepackage{graphicx}
\usepackage{subfigure}
\usepackage{booktabs} 

\usepackage{hyperref}

\usepackage[accepted]{icml2023}

\usepackage{amsmath}
\usepackage{amssymb}
\usepackage{mathtools}
\usepackage{amsthm}
\usepackage{multirow}
\usepackage{enumitem}

\usepackage{xspace}

\usepackage[capitalize,noabbrev]{cleveref}

\theoremstyle{plain}
\newtheorem{theorem}{Theorem}[section]
\newtheorem{proposition}[theorem]{Proposition}

\newtheorem{corollary}[theorem]{Corollary}
\theoremstyle{definition}

\theoremstyle{remark}

\usepackage[textsize=tiny]{todonotes}

\newcommand{\E}[0]{\mathbb{E}}

\newcommand{\indicator}[1]{1\{#1\}}

\newcommand{\abv}[0]{DWSL\xspace}
\newcommand{\fullname}[0]{Distance Weighted Supervised Learning\xspace}

\newif\ifcomments
\commentstrue

\ifcomments
\newcommand{\jh}[1]{{\color{blue} [JH: #1]}}
\newcommand{\ds}[1]{{\color{red} [DS: #1]}}
\newcommand{\jg}[1]{{\color{magenta} [JG: #1]}}

\else
\newcommand{\jh}[1]{}
\newcommand{\ds}[1]{}
\newcommand{\jg}[1]{}

\fi

\icmltitlerunning{\fullname}

\begin{document}

\twocolumn[
\icmltitle{\fullname for Offline Interaction Data}



\icmlsetsymbol{equal}{*}

\begin{icmlauthorlist}
\icmlauthor{Joey Hejna}{stanford}
\icmlauthor{Jensen Gao}{stanford}
\icmlauthor{Dorsa Sadigh}{stanford}
\end{icmlauthorlist}

\icmlaffiliation{stanford}{Department of Computer Science, Stanford University}

\icmlcorrespondingauthor{Joey Hejna}{jhejna@cs.stanford.edu}

\icmlkeywords{Machine Learning, ICML}

\vskip 0.3in
]



\printAffiliationsAndNotice{}  

\begin{abstract}
    Sequential decision making algorithms often struggle to leverage different sources of unstructured offline interaction data. Imitation learning (IL) methods based on supervised learning are robust, but require optimal demonstrations, which are hard to collect. Offline goal-conditioned reinforcement learning (RL) algorithms promise to learn from sub-optimal data, but face optimization challenges especially with high-dimensional data. To bridge the gap between IL and RL, we introduce \fullname or \abv, a supervised method for learning goal-conditioned policies from offline data. \abv models the entire distribution of time-steps between states in offline data with only supervised learning, and uses this distribution to approximate shortest path distances. To extract a policy, we weight actions by their reduction in distance estimates. Theoretically, \abv converges to an optimal policy constrained to the data distribution, an attractive property for offline learning, without any bootstrapping. Across all datasets we test, \abv empirically maintains behavior cloning as a lower bound while still exhibiting policy improvement. In high-dimensional image domains, \abv surpasses the performance of both prior goal-conditioned IL and RL algorithms. Visualizations and code can be found at \url{https://sites.google.com/view/dwsl/home}.
\end{abstract}

\section{Introduction}
\label{sec:intro}

Many advancements in deep learning
are underpinned by a markedly similar formula: collect a large, diverse dataset and train a model with supervised learning. Thus, for sequential decision making problems it is naturally desirable to apply the same formula yet again. This is especially true when learning multi-task policies that operate on high-dimensional image observations, which is often the case for real-world robotics applications. Training generalizable robot policies will likely require scaling dataset size and model capacity, which supervised learning algorithms have excelled at. \looseness=-1

One promising paradigm for scalable robot learning is offline goal-conditioned learning, where policies are trained to reach arbitrary goals from interaction data without dense reward annotation. This can be done via supervised learning with goal-conditioned imitation learning (IL) \citep{ghosh2021learning, emmons2022rvs}, where dataset actions are cloned by a policy. IL methods tend to excel when given access to broad data in the form of expert demonstrations \cite{jang2022bc} or unstructured play \cite{lynch2019play}. However, collecting vast quantities of this type of data has proven to be tedious at best \citep{akgun2012keyframe}, and any sub-optimalities within the data can be detrimentally materialized by the learned policy. \looseness=-1

Value-based offline reinforcement learning (RL) algorithms, on the other hand, promise to address these challenges by learning policies that can improve upon the behavior found in sub-optimal datasets. However, RL methods can be difficult to get working in practice due the optimization challenges of temporal difference learning, known as the \textit{deadly triad}, which consists of function approximation, off-policy learning, and bootstrapping \citep{van2018deep,sutton2018reinforcement}. 
These issues are further exacerbated in offline image-based settings, where additional data cannot be collected to correct modeling errors, and high-dimensional visual observations make stitching together distinct experience difficult. As a result, despite complicated tricks and careful hyperparameter tuning, the performance of offline RL methods can vary dramatically from dataset to dataset. For example, while offline RL algorithms perform well on data collected by other RL agents, they have been known to fail on data collected by humans \citep{robomimic2021} or exploratory strategies \citep{yarats2022exorl}. \looseness=-1 

Ideally, we want scalable algorithms that perform well on the widest variety of datasets. To make progress towards this goal, we propose an offline goal-conditioned algorithm that maintains the optimization robustness of supervised learning, but still performs value estimation to improve upon sub-optimal data. Prior work \citep{AWRPeng19} have attempted to do the same with regression-based objectives, but were not designed for the goal-conditioned setting, and in practice \textit{still used boostrapping}. 


Our key insight is to view goal-conditioned value prediction --- which enables policy improvement beyond imitation --- as \emph{a supervised classification problem}. Instead of learning optimal values directly, we model the entire empirical distribution of discrete distances between states in offline data. Then, we compute meaningful statistics of this distribution to extract shortest path estimates between any two states without the optimization challenges of bootstrapping. Specifically, we use the LogSumExp as a smooth estimate of the minimum distance. Finally, we re-weight actions by how closely they reduce the estimated distance to the goal. We call our approach \fullname or \abv. Unlike prior supervised approaches to RL, \abv theoretically converges to an optimal policy constrained to the behavior distribution of the dataset, instead of doing only a single step of policy improvement. 

Empirically, \abv has several attractive properties that we believe contribute to its practical performance and broad applicability. \abv does not require access to a dense reward function, goal labels, or optimal demonstrations.
We find \abv to be extremely robust to hyperparameters and datasets, which we posit is due to the usage of only supervised learning methods as subroutines. In our experiments, \abv exceeds the performance of imitation learning on 23 of the 27 datasets we test \textit{using the same algorithm-specific hyperparameters}. \abv mathces IL on 3 of the remaining 4. On 4 of 5 human-collected datasets and 6 of 8 image datasets, \abv surpasses the performance of the best offline GCRL baseline. Because \abv inherits the properties of supervised learning while still exhibiting policy improvement, we believe it is a prime candidate for scaling to larger, more realistic datasets comprised of high-dimensional image observations. 


\section{Related Work}

Learning to reach goals is a long studied problem in both imitation learning (IL) and reinforcement learning (RL). IL approaches to goal-reaching often clone the actions of an expert conditioned on a future goal. While IL has shown promising results on expert demonstrations \citep{argall2009survey,rt12022arxiv, duan2017one} and play \citep{lynch2019play, belkhale2022plato, pertsch2021accelerating}, its performance directly depends on dataset optimality. To address these limitations, past works have used additional online data \citep{gupta2020relay, ding2019goal}, which is difficult to collect, or attempt to directly correct for the ability of experts \citep{pmlr-v162-beliaev22a,NEURIPS2021_670e8a43, cao2021learning}.

Other approaches leverage goal-conditioned RL (GCRL) \citep{kaelbling1993learning} to improve upon offline sub-optimal data. GCRL algorithms learn policies with sparse goal-reaching reward functions, which can be difficult to optimize, particularly when there is little data of the agent actually reaching its goal. To make datasets appear more optimal, most GCRL algorithms leverage hindsight relabeling \citep{andrychowicz2017hindsight}, or other strategies \citep{eysenbach2020rewriting, li2020generalized}, where achieved outcomes are designated as desired goals. Nevertheless, traditional GCRL methods \citep{eysenbach2020c, kaelbling1993learning, pmlr-v37-schaul15, eysenbach2019search} are based on dynamic programming, which can be difficult in high-dimensional spaces with function approximation \cite{van2018deep}. 

In the offline setting we consider, these problems are amplified by out-of-distribution (OOD) errors when sampling actions that cannot be corrected with new data. Offline GCRL algorithms have attempted to address OOD errors by being pessimistic about actions outside of the dataset \citep{chebotar2021actionable, rosetelatent, kumar2020conservative}, using weighted imitation learning objectives \citep{ma2022offline,yang2022rethinking}, using model-based planning \citep{tian2020model}, or by estimating implicit quantities \citep{fanggeneralization} to avoid the need for sampling \citep{kostrikov2022offline, garg2023extreme}. Despite these efforts, offline RL methods do not always outperform IL \citep{robomimic2021}.

In between IL and RL, some methods optimize policy improvement objectives using only supervised learning. \citet{ghosh2021learning} do so for goal reaching using only hindsight optimality. In settings with access to reward, \citet{peters2007reinforcement, wang2018exponentially} optimize for a single step of trust region policy improvement \citep{schulman2015trust} using regression. \citet{AWRPeng19} also solves for the same objective with regression, but their practical algorithm still uses bootstrapping. \citet{Hartikainen2020Dynamical} consider GCRL with distances, but do so online and use regression. Unlike all these methods, \abv is designed for offline learning, fits a discrete distribution, and solves for an optimal constrained policy, not a single-step improvement. Other ``upside-down'' approaches to RL condition on reward to replicate the best performance found in offline datasets \cite{emmons2022rvs, srivastava2019training}. However, all of these methods require reward labels and do not guarantee policy improvement. 

Recent works have also scaled existing approaches to larger transformer models. \citet{cui2022play} does so for goal-conditioned IL, Decision Transformer \citep{chen2021decision} does so for upside-down RL, and Trajectory Transformer \citep{janner2021offline} does so for model-based planning. Even though the latter two of are not goal-conditioned, we view all of their general insights as complementary to \abv. While we focus on evaluating \abv's algorithmic approach by holding architectures constant, like the aforementioned transformer-based approaches, \abv uses only supervised objectives, and thus could be combined with similar sequence models in practice. \looseness=-1

\section{\fullname}
\label{sec:method}
We divide the description of our method across three sections. In Sec.~\ref{sec:setup}, we describe the ``learning from offline interaction data'' setting. Next in Sec.~\ref{sec:simple} we provide a high-level overview of \abv in terms of learning distance estimates. In Sec.~\ref{sec:complex}, we then formalize and translate this intuition into the offline goal-conditioned reinforcement learning (GCRL) paradigm.
A full algorithm block for our method can be found in Appendix \ref{app:alg}.

\subsection{Learning from Offline Interaction Data}
\label{sec:setup}
We seek to develop methods that can leverage the broadest set of training data. Consequently, we assume the agent acts in a deterministic Markov Decision Process $\mathcal{M} = (\mathcal{S}, \mathcal{A}, f(s,a), \mathcal{G}, r(s,a,g), \gamma)$ with state space $\mathcal{S}$, action space $\mathcal{A}$, deterministic dynamics $s_{t+1}= f(s_t,a_t)$ where $s_{t+1}$ is the resulting state from taking action $a_t$ at state $s_t$, goal space $\mathcal{G}$, sparse goal-conditioned reward function $r(s,a,g)$, and discount factor $\gamma$. The goal space $\mathcal{G}$ is a subspace of the state space $\mathcal{S}$ admitted by a goal-extraction function $g = \phi(s)$, which is often the identity $\phi(s) = s$. Our objective is to learn a goal-conditioned policy $\pi(a|s,g)$ that has mastery over its environment by being able to reach and remain at goals. To capture this, we  maximize the expected discounted return of a reward function $r(s,a,g)$ given a goal distribution $p(g)$:
\begin{equation}
    \label{eq:vanilla_objective}
    \max_\pi \E_{g \sim p(g), a \sim \pi(\cdot|s, g)} \left[\sum_{t=0}^\infty \gamma^t r(s_t, a_t, g)\right].
\end{equation}
While this exact setup differs from prior work, it shares strong connections with two common problem settings: the Stochastic Shortest Path (SSP) problem \citep{bertsekas1991analysis} and GCRL. If we choose $\phi$ to be the identity, $p(g)$ to be uniform, and the reward function to be $r(s_t, a_t, g) = -\indicator{s_t \ne g}$ under known, stochastic dynamics, we recover the SSP problem. 
However, SSP assumes the ground-truth dynamics to be known, which is not the case when learning purely from offline trajectories.

Alternatively, if we choose $r(s_t, a_t, g) = p(\phi(s_{t+1}) = g | s_t, a_t)$ and have access to a known $p(g)$, we recover GCRL. Note that works in GCRL assume that each trajectory is labeled with the policy's intended goal, providing information about the test-time goal distribution $p(g)$. While seemingly innocuous, this assumption limits the data that offline GCRL can learn from. Many offline data sources, like unstructured play \citep{lynch2019play} or unsupervised exploration \citep{yarats2022exorl}, do not contain goal labels along with each trajectory. Moreover, goals can be hard to obtain. In image domains for example, collecting a goal label would require constructing a scene by hand where the desired task has been solved. \looseness=-1

To learn from the broadest set of offline data, we consider a more general setting where we do not assume access to ground-truth dynamics, reward labels, or the test-time goal distribution a priori.
At training time, we are only given a dataset of state-action trajectories of an arbitrary level of optimality. We denote trajectories of horizon $T$ as $\tau = (s_0, a_0, s_1, a_1, \dots, a_{T-1}, s_T)$, and the entire dataset of trajectories as $\mathcal{D} = \{(\tau)\}$. We take $p(g)$ to be the distribution of goals induced by applying the goal extraction function $\phi$ over all states in the dataset. While we could try to use a uniform distribution like in SSP, we posit that for most practical datasets goals around the data distribution are likely closer to those for tasks of interest. Our method can use any sparse indicator reward function that can be computed purely from state-action sequences, but in practice we find empirically estimating $r(s_t,a_t,g)$ as $-\indicator{\phi(s_{t+1}) \ne g}$ to work well.  \looseness=-1


\begin{figure*}[t]
\centering
\includegraphics[width=\textwidth]{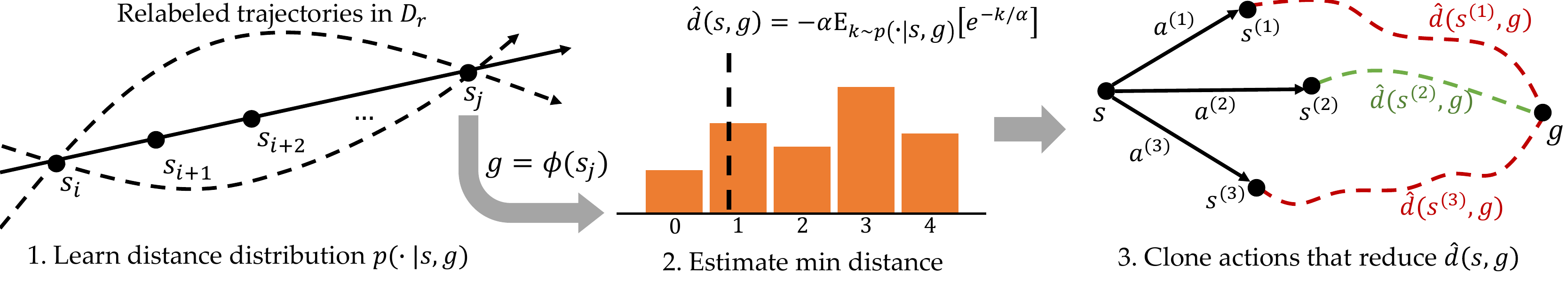}
\vspace{-0.3in}
\caption{A depiction of the three phases of \abv. In the first phase, we learn a distribution of distances over the number of time-steps between states. We then compute the LogSumExp of this distribution in order to estimate the minimum distances to the goal . Finally, we extract a policy by imitating actions that follow reduce the learned distance metric.}
\label{fig:method}

\vspace{-0.2in}
\end{figure*}

\subsection{An Intuitive Explanation of \abv}
\label{sec:simple}
In this section, we describe the \abv algorithm using a distance-based interpretation. Intuitively, the best goal-reaching policy is one that reaches goal $g$ from current state $s$ in the fewest number of time steps, or by the shortest path. However, trajectories within the offline dataset $\mathcal{D}$ do not necessarily follow shortest paths. As a result, imitation learning techniques like GCSL \citep{ghosh2021learning}, which clone the action taken at each state, may exhibit sub-optimal behavior when trained on $\mathcal{D}$. To improve upon imitation policies, prior works in offline GCRL have estimated shortest paths using approximate dynamic programming with a sparse reward function. Such approaches predict a state-action value function $Q^\pi(s,a,g) = \E_\pi [\sum_{t' = t}^\infty \gamma^{t' -t} r(s_t,a_t,g) | s_t = s, a_t = a]$, or the expected discounted return when following policy $\pi$ after taking action $a$ at state $s$. A parameterized $Q$-function, $Q^\pi_\theta$, is then iteratively updated to follow the argmax policy $\pi^*(a|s,g) = \arg \max_a Q_\theta(s,a,g)$ via the following temporal difference (TD) update:
\begin{gather*}
    \min_\theta \E[(Q_\theta(s_t,a_t,g) - y)^2], \\
    y = r(s_t,a_t,g) + \gamma \max_{a_{t+1}} Q_\theta(s_{t+1}, a_{t+1},g).
\end{gather*}
This objective, however, is difficult to optimize in practice. First, TD updates are bootstrapped from next state $Q$-function predictions. Without high coverage of actions $a_t$ and resulting next states $s_{t+1}$, we can end up with inaccurate or even divergent estimates using $Q_\theta$. This is particularly problematic for high-dimensional data like images. Second, the $\max$ operation over $a_{t+1}$ can result in out-of-distribution actions which cause  the $Q$-function to exhibit significant errors. Prior works have used policy constraints \citep{wu2019behavior, fujimoto2021a} and value conservatism \citep{kumar2020conservative} to mitigate this phenomena. More recently, implicit estimation techniques \citep{kostrikov2022offline, garg2023extreme} have been used to eliminate model evaluation on out-of-distribution actions, but still rely on bootstrapping.

\abv eliminates both of these issues. To do so, we 1) only estimate distances with supervised learning and 2) only evaluate our learned models under the support of the data distribution during training. We learn the entire distribution of pairwise distances between states in $\mathcal{D}$, use this distribution to estimate the minimum goal distance contained within the dataset at each state, and then learn a policy to follow these paths. Our entire approach is outlined in Figure \ref{fig:method}.

\noindent \textbf{Learning Distances.} For any two states $s_i, s_j$ in a trajectory $\tau$, where $i < j$, we know that goal $\phi(s_j)$ can be reached from $s_i$ in $j - i$ time steps. Using this \textit{hindsight relabeling} technique \citep{andrychowicz2017hindsight}, we generate a relabeled dataset $\mathcal{D}_r = \{(s_i, a_i, s_{i+1}, \phi(s_j))\}$ that contains all the pairwise distances between states and goals. For a given state and goal pair sampled from $\mathcal{D}_r$, we model the discrete distribution over the number of time-steps $k$ from $s$ to $g$, or $p^r(k|s,g)$ as shown on the left-most side of Figure \ref{fig:method}. We obtain a parameterized estimate $p^r_\theta$ of this distribution via maximum likelihood under the relabeled dataset: 
\begin{equation}
\label{eq:simple_distance}
\max_\theta \E_{\mathcal{D}_r}[\log p^r_\theta(j - i - 1| s_i, \phi(s_j))].
\end{equation}

In practice, $p_\theta^r$ is modeled as a discrete classifier over possible distances. The shortest path between $s$ and $g$ contained within $\mathcal{D}_r$ is then the smallest value of $k$ such that $p^r(k|s,g) > 0$. However, because $p_\theta^r$ is learned using function approximation, estimating the minimum distance in this manner with $p_\theta^r$ will likely exploit modeling errors. Instead, we compute the LogSumExp over the distribution to obtain a soft estimate of the minimum distance:
\begin{equation}
\label{eq:simple_lse}
\hat{d}(s,g) = -\alpha \log \E_{k \sim p^r_\theta(\cdot|s,g)}\left[e^{-k/\alpha}\right].
\end{equation}
Note that we multiply the distances by $-1$ to obtain the minimum estimate, instead of the maximum. $\alpha$ is a temperature hyperparameter such that as $\alpha \xrightarrow{} 0$, $\hat{d}(s,g)$ approaches the minimum distance $k$ that has $p_\theta^r(k|s,g) > 0$.

\noindent \textbf{Leveraging Distances for Policy Learning.} After learning minimum distance estimates, we want to follow the path they induce at each state. For example, assume we are in state $s$ and want to reach goal $g$. We can take one of two actions, $a^{(1)}$ and $a^{(2)}$, which result in states $s^{(1)}$ and $s^{(2)}$ respectively. We would prefer to take action $a^{(1)}$ if $\hat{d}(s^{(1)}, g) < \hat{d}(s^{(2)}, g)$, e.g., it has a smaller estimated distance to the goal. Thus, we want to weight the likelihood of different actions by their resulting distance estimates (right of Figure \ref{fig:method}). However, na\"ively weighting actions this way would apply a larger weighting to all datapoints closer to $g$, as any state far away from $g$ will naturally have a larger distance. We instead weight the likelihood of actions according to their \textit{reduction} in estimated distance to the goal, which we refer to as the advantage. This gives the following imitation learning objective over the relabeled dataset for learning a parameterized policy $\pi_\psi$:
\begin{align}
    \label{eq:simple_imitation}
    &\max_\psi \E_{\mathcal{D}_r}\left[e^{\text{adv}/\beta} \log \pi_\psi(a_i | s_i, \phi(s_j) )\right], \\
    &\text{adv} = \hat{d}(s_i,\phi(s_j)) - 1 - \hat{d}(s_{i+1},\phi(s_j)). \nonumber
\end{align} 
    
If taking action $a_i$ to state $s_{i+1}$ follows the shortest path to $\phi(s_j)$, then $\hat{d}(s_i, \phi(s_j)) = 1 + \hat{d}(s_{i+1}, \phi(s_j))$, and $a_i$ will be given a weight of $1$ in Equation \ref{eq:simple_imitation}. If $a_i$ results in a sub-optimal $s_{i+1}$ that leads to a longer path,  $\hat{d}(s_i, \phi(s_j)) < 1 + \hat{d}(s_{i+1}, \phi(s_j))$, and the assigned weight will be less than $1$. We use exponentiated advantages to ensure all weights are positive, as done in prior work \citep{AWRPeng19, nair2020awac, yang2022rethinking, ma2022offline, kostrikov2022offline, garg2023extreme}, but we learn these advantage weights using only supervised learning. $\beta$ is a temperature hyperparameter that controls weighting.

To summarize, our algorithm first learns $p^r_\theta$ via Equation \ref{eq:simple_distance}, estimates distances according to Equation \ref{eq:simple_lse}, and then learns a policy via Equation \ref{eq:simple_imitation}. In practice we truncated the distance distribution $p^r_\theta$ to predict only over a finite number of bins $B$. As predicting the distribution over a long horizon $T$ may be challenging, we take a similar approach to N-step returns and use $B = T // N$ discrete bins. When the task horizon $T$ is sufficiently small, it suffices to set $B = T$. We normalize the distance values by the number of bins $B$ so that they always lie in $[0, 1]$. 

Following this algorithm theoretically approximates the optimal solution to a constrained version of the objective in Eq. \ref{eq:vanilla_objective}. In the next section, we show this connection by demonstrating that distances are equivalent to expected returns. \looseness=-1

\subsection{Policy Improvement with DWSL}
\label{sec:complex}
In this section, we show that the \abv algorithm theoretically solves for an optimal constrained policy under mild assumptions. Here, we develop a lower bound for the infinite horizon MDPs traditional used in GCRL. In the Appendix, we show that the \abv objective recovers the exact optimal policy for fininte horizon MDPs which we use in practice. We begin by outlining how distances can be translated to the total sum of rewards, or return an agent receives. Using this insight, we show how distances can be substituted into common policy learning objectives. 

\textbf{Equivalence of Distances and Returns.} For sparse reward functions, the number of time-steps it takes a policy to reach a goal can be directly mapped to its return. Specifically, we consider the relabeling policy $\pi_r(a|s,g)$ that produces the relabeled experience present in $\mathcal{D}_r$ and make two assumptions. First, we assume the existence of a stationary action $a^{(s)}$ for each state $s \in \mathcal{S}$ such that $f(s, a^{(s)}) = s$ meaning that once a goal is reached, it is possible to stay there indefinitely. Second, we assume that the relabeling policy $\pi_r(a|s,g)$ always takes the stationary action to stay at an achieved the goal, or $\pi_r(s, \phi(s)) = a^{(s)}$. Because sparse reward functions only depend on whether or not $\phi(s) = g$, a policy's return depends only on how many time-steps it was at the goal. As the relabeling policy $\pi_r$ always remains at the goal after reaching it, its return can be computed from just its time-step distance to the goal. For example, consider the reward function $r(s_t,a_t,g) = -\indicator{\phi(s_{t+1}) \ne g}$ used in prior work in GCRL, which is $0$ if the goal is reached after taking action $a_t$ from $s_t$, and $-1$ otherwise. If $\pi_r$ takes $k$ time steps to reach goal $g$ from state $s_t$, it receives a reward of $-1$ for $k-1$ steps, and then a reward of $0$ forever. Per the geometric series formula, this equates to a total discounted return of $-(1 - \gamma^{k-1})/(1 - \gamma)$. Thus, we establish a one-to-one mapping between the distances and returns of $\pi_r$. This implies that modeling the distribution of time-steps it takes $\pi_r$ to reach goals is equivalent to modeling the distribution of returns. As shown in Section \ref{sec:simple}, the distribution of time-steps $p^r(k|s,g)$ is easily learned in a supervised manner. Next, we show that this distribution over time-steps can be substituted into constrained RL objectives.

\textbf{Bounding KL-Constrained Values}. In this subsection, we present the KL-constrained RL objective and its optimal solution. Then, we show that we can use the distribution of returns of $\pi_r$ to bound the corresponding optimal state-value function. \looseness=-1

KL-constrained RL adds a KL-divergence penalty $D_{KL}(\pi || \pi_r)$ between the learned policy $\pi$ and dataset behavior policy $\pi_r$ to the objective from Equation \ref{eq:vanilla_objective}: \looseness=-1
\begin{equation}
\label{eq:kl_objective}
    \max_\pi \E_{\pi, p(g)}\left[\sum_{t=0}^\infty \gamma^t r(s_t,a_t,g) - \alpha \log \frac{\pi(a_t|s_t,g)}{\pi_r(a_t|s_t,g)}\right].
\end{equation}
In the above, $\alpha$ weights the KL-penalty. Similar objectives have proven to be useful for preventing excessive extrapolation in offline RL \citep{kumar2020conservative}.

As shown by \citet{garg2023extreme,pertsch2021accelerating} using principles first derived in maximum entropy RL \citep{haarnoja2017reinforcement, ziebart2010modeling}, the optimal state value $V^*$ and state-action value $Q^*$ to Equation \ref{eq:kl_objective} satisfy a unique fixed point. In the goal-conditioned setting, this gives
\begin{align}
   & Q^*(s_t,a_t,g) = r(s_t,a_t,g) + \gamma V^*(s_{t+1}, g), \label{eq:kl_q} \\
   & V^*(s_t,g) = \alpha \log \E_{a_t \sim \pi_r(\cdot|s_t,g)}\left[e^{Q^*(s_t,a_t,g)/\alpha}\right] .\label{eq:kl_v}
\end{align}
Notice how the value function $V^*$ is written in terms of the expectation under $\pi_r$. By repeatedly substituting Equation \ref{eq:kl_q} into \ref{eq:kl_v} we remove the dependence on $Q^*$, which we do not know, and obtain an expectation as a function of the discounted returns of $\pi_r$. This is summarized by the following Proposition, for which the full proof is in Appendix \ref{proof:lemma}.

\begin{proposition}
\label{lemma:empirical}
Assuming deterministic dynamics, we can bound the optimal value function for the objective in Equation \ref{eq:kl_objective} using empirical returns of $\pi_r$ as
\begin{equation}
    V^*(s_t,g) \geq \alpha \log \E_{\pi_r}[e^{(\sum_{t'=t}^\infty \gamma^{t' -t} r(s_{t'}, a_{t'},g))/\alpha}].
\end{equation}
\end{proposition}

The lower bound is a result of applying Jensen's inequality to the exponentiated discount factor, $\E[X]^\gamma \geq E[X^\gamma]$. As $\gamma \xrightarrow{} 1$, as is often the case for RL in the real-world, this bound becomes tighter. In Appendix \ref{proof:finite_lemma} we show that this relationship holds with equality for finite horizon MDPs. \looseness=-1

The right hand side is simply the LogSumExp over the discounted returns of $\pi_r$. As the LogSumExp converges to the maximum, we interpret this result as stating that the optimal value function is lower bounded by the best behavior exhibited by $\pi_r$. As shown before, we can equate the distribution over discounted returns with the distribution over time-step distances.
Thus, via a simple change of variables, we arrive at the following corollary.

\begin{corollary}
\label{corollary:distance}
Assume there exists an onto mapping from discounted returns to time-step distances $k$ such that $\sum_{t=0}^\infty \gamma^t r(s_t,a_t,g) = \mathcal{R}_k$ for some $k \in \mathbb{N}$, then 
\begin{equation}
    V^*(s_t,g) \geq \alpha \log \E_{k \sim p^r(\cdot |s_t, g)} \left[e^{\mathcal{R}_k / \alpha}\right].
\end{equation}
\end{corollary}

The full proof is in Appendix \ref{proof:corollary}. Notice that the form of this bound is identical to the distance estimator used in Equation \eqref{eq:simple_distance} when $r(s_t,a_t,g) = -\indicator{\phi(s_{t+1}) = g}$, implying that the distance learning component of \abv does indeed extract optimal value function estimates from just the distance distribution $p^r(\cdot|s,g)$. While most works depend on bootstrapping to estimate the optimal value function, we have shown that we can obtain similar estimates using only supervised learning by taking advantage of the structure of reward functions in offline GCRL. 

\textbf{Policy Extraction.} The final step of the \abv algorithm is to extract the optimal policy from distance, or value, estimates. Given the optimal $Q^*$ and $V^*$, \citet{AWRPeng19} showed that the optimal policy $\pi^*$ for the KL-constrained RL objective can be written in proportion to the advantage function $A^*(s, a, g) = Q^*(s,a,g) - V^*(s, g)$ and the behavior distribution $\pi_r$. Given that we assume deterministic dynamics, we can remove the dependence of $A^*$ on $Q^*$ via Equation \ref{eq:kl_q}, and arrive at the following statement:
\begin{align}
    \pi^*(a_t|s_t,g) &\propto \pi_r(a_t|s_t,g) \exp \left( A^*(s_t, a_t, g) /\alpha \right), \\
    A^*(s_t, a_t, g) &= r(s_t,a_t,g) + \gamma V^*(s_{t+1},g) - V^*(s_t,g). \nonumber
\end{align}
As is common practice \cite{nair2020awac}, we project this result to the parameterized space of a learned policy $\pi_\psi$ by solving $\psi^* = \arg \min_\psi E_{\mathcal{D}_r} [D_{KL}(\pi^* || \pi_\psi)]$. Doing so exactly recovers the policy learning objective in Equation \ref{eq:simple_imitation} in Section \ref{sec:simple} when using the reward function  $r(s_t,a_t,g) = -\indicator{\phi(s_{t+1}) = g}$. Thus, in the finite horizon case where Lemma \ref{lemma:empirical} is tight (see Appendix \ref{proof:finite_lemma}), \abv recovers the optimal policy corresponding to the objective in Equation \ref{eq:kl_objective}. 


We note that when using the expectation instead of the LogSumExp, \abv recovers the advantage-weighted regression (AWR) objective \citep{AWRPeng19} and corresponds to one-step of KL-constrained policy improvement. We provide derivations of this connection in Appendix \ref{app:awr} and ablate our choice of the LogSumExp statistic in Sec \ref{sec:objective}.

\section{Experiments}
\label{sec:experiments}
In this section we seek to answer the following questions: 1) How does \abv perform across a broad selection of robotic datasets? 2) Does \abv use the right objective? 3) When does \abv fail? 4) How robust  is \abv?

\begin{figure}
\includegraphics[width=0.9\columnwidth]{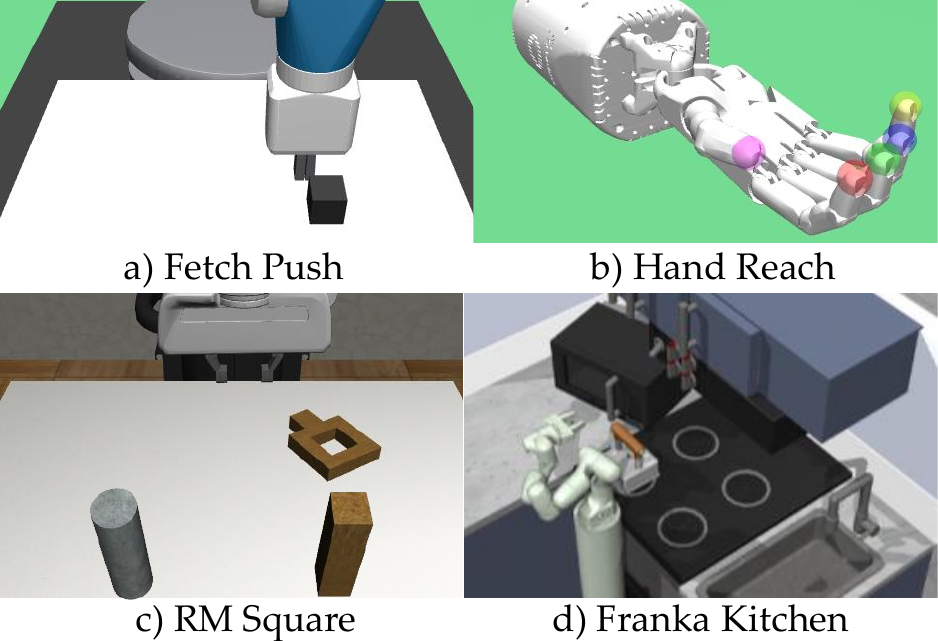}
\centering
\vspace{-0.1in}
\caption{Environment depictions. For Hand and Fetch we show (cropped) image observations provided to agents.}
\label{fig:envs}
\vspace{-0.27in}
\end{figure}

\begin{table*}[t]
\resizebox{\textwidth}{!}{
\centering
\begin{tabular}{cccccccc}
\textbf{Environment}               & \textbf{Dataset} & \textbf{GCSL}           & \textbf{WGCSL}  & \textbf{GoFAR}           & \textbf{GCIQL}  & \textbf{DWSL-B}         & \textbf{DWSL}           \\ \hline
\textbf{Franka Kitchen}                     & \multicolumn{1}{l|}{566 Play Demos}   & \textbf{2.98 $\pm$ .10} & 2.62 $\pm$ .19  & 2.43 $\pm$ .06           & 2.81 $\pm$ .19  & \textbf{2.97 $\pm$ .16} & \textbf{2.92 $\pm$ .04} \\[0.2cm]
\multirow{2}{*}{\shortstack{\textbf{RM Can} \\ \includegraphics[width=0.5in,page=5]{figures/icons.pdf}}}            & \multicolumn{1}{l|}{100 PH Demos}     & 67 $\pm$ 17\%           & 50 $\pm$ 19\%   & 44 $\pm$ 8\%             & 58 $\pm$ 16\%   & 70 $\pm$ 9\%            & \textbf{77 $\pm$ 5\%}   \\
                                   & \multicolumn{1}{l|}{300 MH Demos}     & 31 $\pm$ 13\%           & 27 $\pm$ 6\%    & 35 $\pm$ 2\%             & 28 $\pm$ 20\%   & \textbf{38 $\pm$ 10\%}  & 36 $\pm$ 8\%            \\[0.2cm]
\multirow{2}{*}{\shortstack{\textbf{RM Square} \\ \includegraphics[width=0.5in,page=6]{figures/icons.pdf}}}         & \multicolumn{1}{l|}{100 PH Demos}     & \textbf{48 $\pm$ 9\%}   & 4 $\pm$ 4\%     & 18 $\pm$ 5\%             & 44 $\pm$ 4\%    & 38 $\pm$ 3\%            & \textbf{47 $\pm$ 12\%}  \\
                                   & \multicolumn{1}{l|}{300 MH Demos}     & 10 $\pm$ 5\%            & 7 $\pm$ 3\%     & \textbf{17 $\pm$ 7\%}    & 13 $\pm$ 4\%    & 15 $\pm$ 9\%            & 14 $\pm$ 5\%            \\[0.2cm]
\multirow{2}{*}{\shortstack{\textbf{Fetch Push} \\ \includegraphics[width=0.5in,page=1]{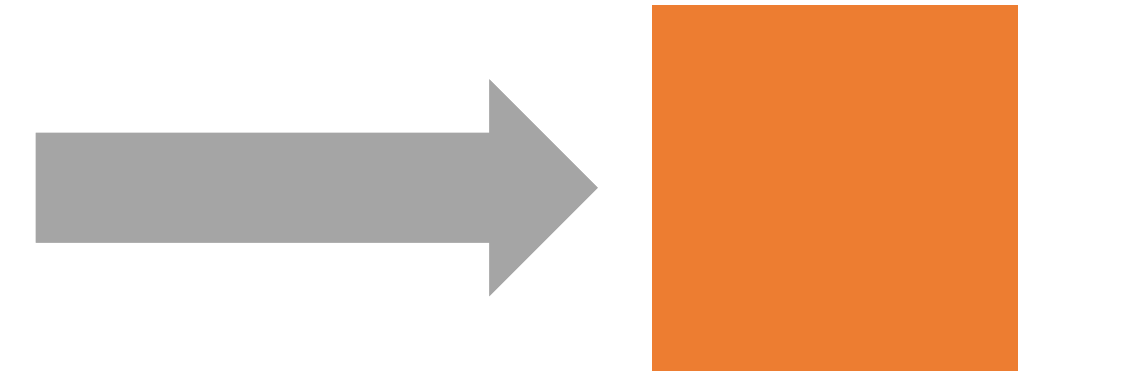}}}  & \multicolumn{1}{l|}{250K Noise 1}     & 18.53 $\pm$ .5          & 22.94 $\pm$ .4  & 14.15 $\pm$ 1.8          & 20.62 $\pm$ .5  & \textbf{24.78 $\pm$ .4} & 22.88 $\pm$ .6          \\
                                   & \multicolumn{1}{l|}{250K Noise 2}     & 10.29 $\pm$ .2          & 12.89 $\pm$ .4  & 7.34 $\pm$ .4            & 11.76 $\pm$ .4  & 15.23 $\pm$ .7          & \textbf{16.57 $\pm$ .2} \\[0.2cm] 
\multirow{2}{*}{\shortstack{\textbf{Fetch Pick} \\ \includegraphics[width=0.5in,page=2]{figures/icons.pdf}}}  & \multicolumn{1}{l|}{250K Noise 1}     & 26.55 $\pm$ 1.1         & 27.97 $\pm$ .7  & 26.23 $\pm$ .5           & 25.33 $\pm$ 1.2 & \textbf{29.86 $\pm$ .1} & \textbf{30.46 $\pm$ .2} \\
                                   & \multicolumn{1}{l|}{250K Noise 2}     & 11.65 $\pm$ .6          & 12.32 $\pm$ 1.2 & 10.32 $\pm$ .3           & 11.08 $\pm$ .5  & 14.26 $\pm$ .3          & \textbf{17.45 $\pm$ .4} \\[0.2cm]
\multirow{2}{*}{\shortstack{\textbf{Fetch Slide} \\ \includegraphics[width=0.5in,page=3]{figures/icons.pdf}}} & \multicolumn{1}{l|}{250K Noise 0.5}   & \textbf{3.61 $\pm$ .2}  & 3.08 $\pm$ .1   & 2.19 $\pm$ .2            & 3.46 $\pm$ .2   & \textbf{3.53 $\pm$ .1}  & \textbf{3.67 $\pm$ .1}  \\
                                   & \multicolumn{1}{l|}{250K Noise 1}     & 2.23 $\pm$ .1           & 2.02 $\pm$ .1   & 1.55 $\pm$ .1            & 2.40 $\pm$ .2   & 2.44 $\pm$ .2           & \textbf{2.60 $\pm$ .1}  \\[0.2cm]
\multirow{2}{*}{\shortstack{\textbf{Hand Reach} \\ \includegraphics[width=0.5in,page=4]{figures/icons.pdf}}}  & \multicolumn{1}{l|}{1M 90\% R 10\% E} & 1.74 $\pm$ 1.7          & 5.41 $\pm$ 3.5  & 5.99 $\pm$ 1.4           & 3.97 $\pm$ 3.8  & 5.91 $\pm$ 1.9          & \textbf{9.53 $\pm$ 3.8} \\
                                   & \multicolumn{1}{l|}{500K Noise 0.2}   & 4.07 $\pm$ 2.0          & 9.46 $\pm$ 3.7  & \textbf{16.70 $\pm$ 2.3} & 5.40 $\pm$ 2.4  & 10.49 $\pm$ 2.2         & 10.79 $\pm$ 1.2        
\end{tabular}
}
\vspace{-0.04in}
\caption{This table includes results across the primary environments we test. Bolded numbers are within 95\% of the best score. We run four seeds in state-based domains and three seeds in image domains. The dataset column provides details on the type of demonstrations. In Robomimic (RM), ``PH'' indicates that the demos came from a single proficient user and ``MH'' indicates that the demos came from multiple users of varying abilities. For Gym Image datasets, the first value gives the number of transitions in the dataset. For datasets with ``Noise'', we provide the standard deviation of the Gaussian noise applied to the oracle policy when collecting the dataset. ``90\% R 10\% E'' refers to 90\% random, 10\% expert data.}
\label{tab:results}
\vspace{-0.2in}
\end{table*}

\subsection{How does \abv Perform?}
We extensively evaluate the performance of \abv on offline interaction datasets across a variety of simulated robotics environments. In this section we present results on the following domains, but include more details, additional results, and learning curves in Appendix \ref{app:experiments}.  

\textbf{Visual Gym Robotics}. The Gym robotics environments from \citet{DBLP:journals/corr/abs-1802-09464}, including Fetch and Hand, are designed to test GCRL algorithms. We construct image-based offline datasets by applying a large amount of action noise to ground-truth policies trained with TD3-HER as in \citet{yang2022rethinking}. We use these datasets to evaluate learning from high-dimensional, sub-optimal data. \looseness=-1

\textbf{Franka Kitchen}. The Franka Kitchen dataset from \citet{lynch2019play} is comprised of 566 human demonstrations and tests how well different methods can learn from play data. At test time, we sample goals from the end of a held-out set of validation trajectories and measure how many of four sequential tasks are completed.

\textbf{Robomimic}. We take the \textit{Square} and \textit{Can} datasets from \citet{robomimic2021} with both proficient and multi-human demonstrations and make them goal-conditioned. At test time, we condition on success states from the validation set. RL algorithms in the past have exhibited poor performance on this benchmark. Consequently, these datasets test \abv's ability to maintain IL as a lower bound. 

We compare the performance of \abv to baselines suitable for learning goal-conditioned policies from offline interaction data without reward or goal labels. GCSL \citet{ghosh2021learning} is an imitation learning method that leverages hindsight relabeling. We also compare to state-of-the-art offline GCRL algorithms that perform TD-learning. WGCSL \citep{yang2022rethinking} uses Q-Learning to estimate shortest paths. GoFAR \citep{ma2022offline} performs value iteration over a dual form of GCRL. GCIQL is a goal conditioned version of IQL \citep{kostrikov2022offline} which uses expectile regression to avoid sampling OOD actions. Finally, we compare to a variant of our algorithm that uses bootstrapping via distributional RL to fit the distance distribution, which we denote DWSL-B. For our method, we use the same hyperparameters $(\alpha, \beta)$ for every single dataset. We vary the number of bins $B$ with the task horizon. Our full results can be found in Table \ref{tab:results}. \looseness=-1

\begin{figure*}[t]
\begin{minipage}[b]{0.49\linewidth}
\centering
\includegraphics[width=\textwidth]{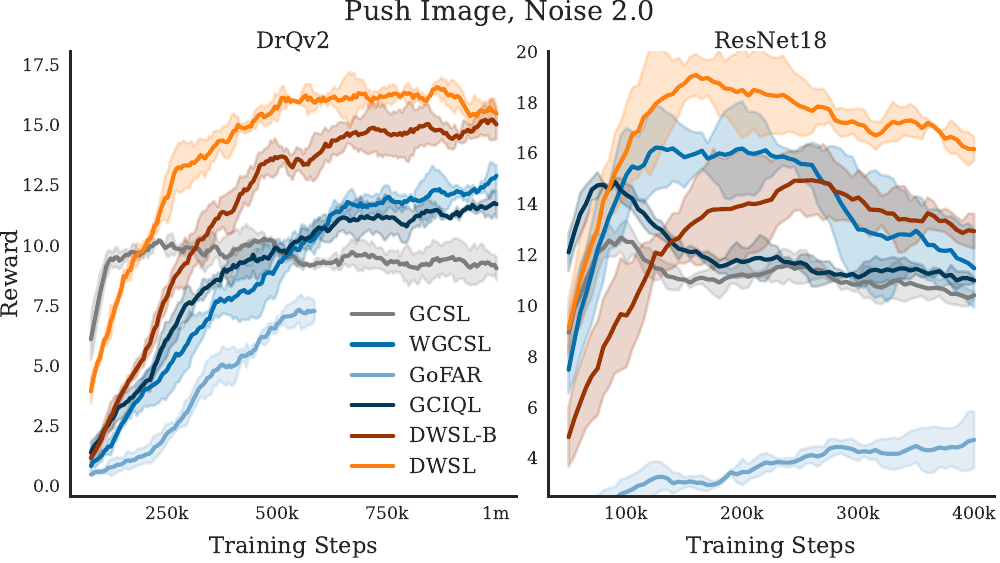}
\end{minipage}
\hspace{0.08cm}
\begin{minipage}[b]{0.49\linewidth}
\centering
\includegraphics[width=\textwidth]{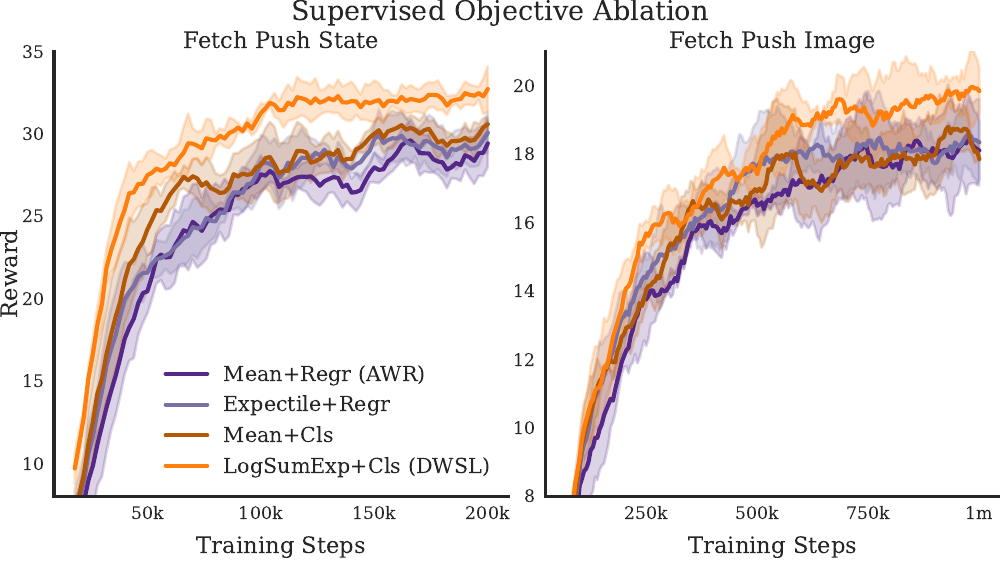}
\end{minipage}
\vspace{-0.1in}
\caption{Here we present ablation results on Fetch Push. From left to right, the first two plots show learning curves for image datasets with different encoders. The first encoder is adapted from DrQv2 \citep{yarats2021improving} (used in main experiments), and the second encoder is adapted from \citet{robomimic2021} and uses a ResNet-18. The right two plots show the performance of different supervised objectives on state and image datasets with 90\% random and 10\% expert data. ``Cls'' stands for classification and ``Regr'' for regression. Our LogSumExp statistic is better able to separate expert from random data than objectives proposed by prior work, like \citet{AWRPeng19}.}
\label{fig:ablation}
\vspace{-0.05in}
\end{figure*}

Overall, we find that \abv surpasses or matches the performance of our supervised baseline, GCSL, on \textit{all} of these datasets. This is not the case for any offline GCRL methods. Moreover, \abv outperforms prior TD-learning based offline GCRL methods on Franka Kitchen and 3 of 4 Robomimic datasets, supporting past evidence \citep{robomimic2021} that TD-learning often struggles with play or human collected data. Notably, TD-learning approaches also perform significantly worse in image domains, despite the fact that the data comes from an oracle RL policy. \abv outperforms or matches all offline GCRL algorithms on 6 of 8 image-based Gym Robotics datasets, while converging faster. We show a sample learning curve in Figure \ref{fig:ablation}. The gap between \abv and bootstrapping approaches is largest in the more sub-optimal, higher noise image datasets. This is potentially because dynamic programming in high-dimensions is harder with sparser coverage. Offline GCRL algorithms beat \abv only in the low-dimensional state-based Gym Robotics domains, but \abv still exhibits improvement over pure imitation. These results are provided in Appendix \ref{sec:experiments}. Finally, \abv outperforms or matches DWSL-B on 7 of 8 image-based Gym Robotics datasets, while performing similarly in other domains, suggesting that with our distributional learning formulation and LogSumExp distance estimation, bootstrapping is not critical for performance. \looseness=-1


\subsection{Does \abv use the right objective?}
\label{sec:objective}
Unlike other approaches, \abv fits the KL-constrained optimal policy by learning a discrete distribution and using the LogSumExp for distance estimation. We analyze if this is the correct choice by comparing \abv to other supervised variants that use different objectives. In Figure \ref{fig:ablation}, we compare \abv's use of the LogSumExp and classification objective to other options in both state and image domains. We first compare to replacing the LogSumExp with the expectation. We also compare to using continuous regression rather than classification, for modeling the expectation as proposed by AWR \citep{AWRPeng19}, and modeling expectiles as proposed by \citep{kostrikov2022offline}. To exaggerate the ability of these approaches to improve behavior, we use datasets comprised of 90\% random and 10\% expert trajectories and for the state dataset, decreased $\alpha$. Overall, we find that our method performs the best, suggesting that our decision to model entire discrete distributions and use the LogSumExp for distance estimation are important for best performance when only using supervised learning for advantage estimation. \looseness=-1

\subsection{When does \abv fail?}
\label{sec:fail}
We run experiments on a goal-conditioned version of the AntMaze benchmark from \citet{fu2020d4rl} to find failure modes of \abv. This toy benchmark requires value propagation to stitch together different parts of sub-optimal experience present in the data. We expect \abv to perform worse on stitching, because it does not use approximate dynamic programming. Instead, \abv can be thought of as performing greedy search for the best action supported by the dataset at each state. Here, \abv consistently improves upon GCSL and other supervised methods (results in Appendix \ref{app:experiments}) but performs worse than GoFAR and GCIQL. This indicates that a lack of bootstrapping is a fundamental drawback to all supervised algorithms. When dealing with more realistic high-dimensional robotics data, however, we believe the benefits of supervised learning can outweigh this downside. \looseness=-1


\begin{table}[]
\centering
\resizebox{\columnwidth}{!}{
\begin{tabular}{lccccc}
\textbf{Dataset} & \textbf{GCSL} & \textbf{WGCSL} & \textbf{GoFar}      & \textbf{GCIQL}      & \textbf{DWSL} \\ \hline
Umaze            & 64 $\pm$ 2    & 83 $\pm$ 3     & \textbf{91 $\pm$ 1} & 85 $\pm$ 1          & 74 $\pm$ 4    \\
Umaze Diverse    & 59 $\pm$ 1    & 47 $\pm$ 6     & \textbf{86 $\pm$ 3} & \textbf{86 $\pm$ 2} & 67 $\pm$ 3    \\
Med Play         & 56 $\pm$ 6    & 35 $\pm$ 22    & 70 $\pm$ 1          & \textbf{74 $\pm$ 5} & 69 $\pm$ 8    \\
Med Diverse      & 60 $\pm$ 3    & 27 $\pm$ 13    & 63 $\pm$ 4          & \textbf{72 $\pm$ 7} & 68 $\pm$ 6    \\
Large Play       & 17 $\pm$ 5    & 0 $\pm$ 0      & \textbf{40 $\pm$ 7} & 31 $\pm$ 8          & 15 $\pm$ 5    \\
Large Diverse    & 12 $\pm$ 3    & 3 $\pm$ 3      & \textbf{45 $\pm$ 8} & 22 $\pm$ 4          & 18 $\pm$ 2   
\end{tabular}
}
\vspace{-0.1in}
\caption{Success rates on the AntMaze-v2 benchmark from \citet{fu2020d4rl}. Learning curves in Appendix \ref{app:experiments}.}
\label{tab:antmaze}
\vspace{-0.25in}
\end{table}

\subsection{How robust is \abv?}
In this section, we investigate how the performance of \abv is affected by changes in architecture, augmentation, hyperparameters, and access to ground-truth goals.

To test how well \abv scales, we run experiments on the Gym Robotics image datasets with ResNet-18 architectures from \citet{robomimic2021}. An example learning curve on the ``Fetch Push Noise 2.0'' dataset can be found in Figure \ref{fig:ablation}, and all learning curves can be found in Appendix \ref{app:ablations}. Overall performance increases for most methods with this additional network capacity, though overfitting starts earlier. Interestingly, the gap between \abv and DWSL-B increases with the ResNet architecture, which we believe indicates that \abv's supervised objectives are more capable of scaling with model size. We train encoders with gradients from the policy instead of the value function because it performed better for all methods (Appendix \ref{app:ablations}), suggesting that supervised learning leads to stronger representation learning in high-dimensions.

DWSL's supervised value function function is also more robust to image-based perturbations than dynamic programming based GCRL algorithms. In Appendix \ref{app:ablations}, we measure the pearson correlation of estimated values on expert demonstrations in Fetch Push with and without random shift augmentations. \abv's pearson correlation coefficient drops by only 0.006 under augmentation, while other methods can drop by more than ten times that amount.

We test the sensitivity of \abv to different values of $\alpha$ and $\beta$ in state domains in Appendix \ref{app:ablations}. Varying $\alpha$ from $0.1$ to $10$ only changes return by $0.2$ and $2.86$ on Fetch Push state with noisy expert and 90\% random data respectively. Varying $\beta$ from $0.01$ to $0.25$ changes return by $5.87$ and $1.52$ on the same benchmarks. This is unlike many offline RL algorithms, where temperature parameters have been shown to drastically affect performance \citep{garg2023extreme}.\looseness=-1

Finally, in Appendix \ref{app:ablations} we consider how performance might change if we give algorithms access to the test-time goal distribution $p(g)$ by testing \abv and WGCSL with different goal relabeling ratios, which controls how often goals are relabeled with achieved states during training. When the relabeling ratio is 1, we recover the \textit{learning from offline interaction} setting we consider. We find that the performance of WGCSL is affected dramatically when the relabeling ratio is 0.5 -- it substantially degrades in Fetch, but improves in AntMaze -- while \abv is largely unaffected. 
\section{Conclusion}
We propose \fullname, a general method for learning goal-conditioned policies from offline interaction data. Theoretically, we show that by exchanging continuous value estimates with statistics of the empirical discrete distance distribution, we can fit an optimal constrained policy using only supervised learning. Empirically, we demonstrate that \abv outperforms prior work on image domains and unlike previous offline GCRL methods always performs the same or better than IL baselines. This makes \abv suitable for scaling to large, real-world applications with heterogeneous and high-dimensional data. Future work can apply \abv to larger architectures and datasets and explore avenues for pretraining the learned distance distribution on action-free datasets such as internet video. \looseness=-1

\newpage
\section*{Acknowledgements}
This work was supported by ONR, DARPA YFA, Ford, and NSF Awards \#1941722 and \#2218760. JH is graciously supoported by the National Defense Science Engineering Graduate (NDSEG) Fellowship Program. We would also like to thank members of the ILIAD lab and our anonymous reviewers for fruitful discussions. 

\bibliography{references}
\bibliographystyle{icml2023}

\newpage
\appendix
\onecolumn
\section*{Appendix}
We divide the Appendix into four different sections as follows.
\begin{enumerate}[label=\Alph*.]
    \item In Appendix \ref{app:theory} we provide full derivations of mathematical results from \ref{sec:method} and additional theoretical results.
    \item In Appendix \ref{app:alg} we provide a detailed algorithm block for \abv. 
    \item Appendix \ref{app:alg} serves as an extended results section, where we provide full learning curves for experiments and ablations mentioned in Section \ref{sec:experiments}.
    \item Finally, in Appendix \ref{sec:details} we provide hyperparameter and implementation details.
\end{enumerate}

\section{Theory}
\label{app:theory}

\subsection{Proof of Proposition \ref{lemma:empirical}}
\label{proof:lemma}
Assume deterministic dynamics and discount factor $\gamma < 1$. \citet{garg2023extreme} gives us that for objective \ref{eq:kl_objective}, the following equations hold for optimal value functions:
\begin{align*}
    Q^*(s_t,a_t,g) &= r(s_t,a_t,g) + \gamma V^*(s_{t+1}, g) \\
    V^*(s_t, g)  &= \alpha \log \E_{a_t \sim \pi_r(\cdot|s_t, g)}\left[e^{Q^*(s_t,a_t) / \alpha }\right]
\end{align*}

We proceed by taking the optimal value function, substituting in the optimal Q-value, and subsequently the next-timestep value definition. Then, we apply Jensen's Inequality and the Markov property to compute the expectation over both $a_t$ and $a_{t+1}$. Repeating this process recursively, we attain the sum of discounted returns.

\begin{align*}
    V^*(s_t, g)  &= \alpha \log \E_{a_t \sim \pi_r(\cdot|s_t, g)}\left[e^{Q^*(s_t,a_t) / \alpha }\right] \\
    &= \alpha \log \E_{a_t \sim \pi_r(\cdot|s_t, g)}\left[e^{(r(s_t,a_t,g) + \gamma V^*(s_{t+1}, g) )/ \alpha }\right]  \; \text{substitute $Q^*$} \\
    &= \alpha \log \E_{a_t \sim \pi_r(\cdot|s_t, g)}\left[e^{r(s_t,a_t,g)/\alpha}\left(e^{V^*(s_{t+1}, g)/ \alpha}\right)^\gamma \right] \; \text{isolate $V^*$} \\
    &= \alpha \log \E_{a_t \sim \pi_r(\cdot|s_t, g)}\left[e^{r(s_t,a_t,g)/\alpha}\left( \E_{a_{t+1} \sim \pi_r(\cdot|s_{t+1}, g)}\left[e^{Q^*(s_{t+1}, a_{t+1})/\alpha}\right]\right)^\gamma \right] \; \text{substitute $V^*$} \\
    &\geq \alpha \log \E_{a_t \sim \pi_r(\cdot|s_t, g)}\left[e^{r(s_t,a_t,g)/\alpha} \E_{a_{t+1} \sim \pi_r(\cdot|s_{t+1}, g)}\left[e^{\gamma Q^*(s_{t+1}, a_{t+1})/\alpha}\right] \right] \; \text{apply Jensen's to $\gamma$} \\
    &= \alpha \log \E_{a_t, a_{t+1} \sim \pi_r}\left[e^{r(s_t,a_t,g)/\alpha} e^{\gamma Q^*(s_{t+1}, a_{t+1})/\alpha}\right] \; \text{apply the Markov Property}
\end{align*}
We then notice that we have written the value function for time $t$ in terms of the Q function for time $t+1$. We can repeatedly apply the same steps, shown once and then infinitely below, to obtain the final result.
\begin{align*}
 V^*(s_t, g)  &\geq \alpha \log \E_{a_t, a_{t+1} \sim \pi_r}\left[e^{r(s_t,a_t,g)/\alpha} e^{\gamma Q^*(s_{t+1}, a_{t+1})/\alpha}\right] \\
    &= \alpha \log \E_{a_t, a_{t+1}, a_{t+2} \sim \pi_r}\left[e^{r(s_t,a_t,g)/\alpha} e^{\gamma r(s_{t+1}, a_{t+1})/\alpha}e^{\gamma^2 Q(s_{t+2}, a_{t+2})/\alpha}\right] \; \text{repeat once} \\
    &= \alpha \log \E_{\tau \sim \pi_r}\left[e^{r(s_t,a_t,g)/\alpha} e^{\gamma r(s_{t+1}, a_{t+1})/\alpha}e^{\gamma^2 r(s_{t+2}, a_{t+2})/\alpha} ... \right] \; \text{repeat infinitely} \\
    &= \alpha \log \E_{\tau \sim \pi_r}\left[e^{\left(\sum_{t'=t}^\infty \gamma^{t - t'} r(s_{t'},a_{t'},g)\right)/\alpha} \right] \; \text{reduce}
\end{align*}
Note that the only step that is not true with equality was the application of Jensen's Inequality to the discount factor $\gamma$, as $\E[X]^\gamma \geq E[X^\gamma]$. This bound becomes tight as the function is near linear for $\gamma \xrightarrow{} 1$. 

\subsection{Proof of Proposition \ref{lemma:empirical} for Finite Horizons}
\label{proof:finite_lemma}

The bound \abv computes on the optimal KL-constrained policy is tight in the finite horizon case with $\gamma = 1$ which we use in practice. Here we present an analogous proof of Proposition \ref{lemma:empirical} for the finite case. Consider the optimal finite horizon fixed point equations, indexed by time $t$:
\begin{align*}
    Q^*_t(s_t,a_t,g) &= r(s_t,a_t,g) + V^*_{t+1}(s_{t+1}, g) \\
    V^*_t(s_t, g)  &= \alpha \log \E_{a_t \sim \pi_r(\cdot|s_t, g)}\left[e^{Q^*_t(s_t,a_t) / \alpha }\right]
\end{align*}
Note that we have removed $\gamma$ as $\gamma = 1$ for the finite horizon case. We can proceed to follow the same substitution steps. We can then follow the same steps as in Proposition \ref{lemma:empirical}. Below we do so, skipping a few steps as the log is the same:
\begin{align*}
     V^*(s_t, g)  &= \alpha \log \E_{a_t \sim \pi_r(\cdot|s_t, g)}\left[e^{(r(s_t,a_t,g) + V^*_{t+1}(s_{t+1}, g) )/ \alpha }\right] \\
     &= \alpha \log \E_{a_t \sim \pi_r(\cdot|s_t, g)}\left[e^{r(s_t,a_t,g)/\alpha} \E_{a_{t+1} \sim \pi_r(\cdot|s_{t+1}, g)}\left[e^{Q^*_{t+1}(s_{t+1}, a_{t+1})/\alpha}\right]\right] \; \text{substitute $V^*$} \\
     &= \alpha \log \E_{a_t, a_{t+1} \sim \pi_r}\left[e^{r(s_t,a_t,g)/\alpha} e^{Q^*_{t+1}(s_{t+1}, a_{t+1})/\alpha}\right] \\
     &= \alpha \log \E_{\tau \sim \pi_r}\left[e^{\left(\sum_{t'=t}^T r(s_{t'},a_{t'},g)\right)/\alpha} \right]
\end{align*}
In the final line we repeat the the first three steps up to the max horizon $T$.

\subsection{Proof of Corollary \ref{corollary:distance}}
\label{proof:corollary}

Corollary \ref{corollary:distance} substitutes the distribution over trajectories from $\pi_r$ with the distribution of distances $p^{r}(\cdot|s_t,g)$ by assuming there exists a mapping between possible returns and a function of the distance $k$. Specifically,  $\sum_{t=0}^\infty \gamma^t r(s,a,g) = \mathcal{R}_k$ for some $k \in \mathbb{N}$. This is possible for reward functions $r(s,a,g) = -\indicator{\phi(s')  = g}$ because  we assume $\pi_r$ to stay at goals after reaching them. Thus, the valid discounted returns of $\pi_r$ directly correspond to distances.  In the below derivation, we use summations to expand the expectation, but they could be exchanged for integrals. We start with Proposition \ref{lemma:empirical} and expand the expectation.

\begin{align*}
    V^*(s_t, g)  &\geq \alpha \log \E_{\tau \sim \pi_r}\left[e^{\left(\sum_{t'=t}^\infty \gamma^{t - t'} r(s_{t'},a_{t'},g)\right)/\alpha} \right] \\ 
    &= \alpha \log \sum_\tau \pi_r(\tau| s_t, g) e^{\left(\sum_{t'=t}^\infty \gamma^{t - t'} r(s_{t'},a_{t'},g)\right)/\alpha}\; \text{Expand expectation over trajectories} \\
    &= \alpha \log \sum_\tau \sum_k \pi_r(\tau, \phi(s_{t+k}) = g| s_t, g) e^{\left(\sum_{t'=t}^\infty \gamma^{t - t'} r(s_{t'},a_{t'},g)\right)/\alpha}\; \text{law of total probability} \\
    &= \alpha \log \sum_\tau \sum_k \pi_r(\tau, \phi(s_{t+k}) = g| s_t, g) e^{\mathcal{R}_k/\alpha}\; \text{sub in $\mathcal{R}_k$} \\
    &= \alpha \log  \sum_k  e^{-\frac{1 - \gamma^k}{(1 - \gamma)\alpha}} \sum_\tau \pi_r(\tau, \phi(s_{t+k}) = g| s_t, g) \; \text{group like terms, swap sum} \\
    &= \alpha \log  \sum_k  e^{-\frac{1 - \gamma^k}{(1 - \gamma)\alpha}} p^{\pi_r}(k| s_t, g) \; \text{marginalize to distance distribution} \\
    &= \alpha \log  \E_{k \sim p^{\pi_r}(\cdot| s_t, g)} \left[e^{-\frac{1 - \gamma^k}{(1 - \gamma)\alpha}} \right] \; \text{re-write expectation.}
\end{align*}
This corollary also holds for the finite horizon case, by assuming $k \in 0, 1, ... T$.

\subsection{From \abv to AWR}
\label{app:awr}
When using the expectation instead of the LogSumExp, \abv solves for the same objective as AWR \citep{AWRPeng19}, which is similar to \citet{wang2018exponentially}. We show this below. The key theoretical difference between \abv and our approach is that AWR exponentially weights the learned policy by the behavior policy's advantage function, instead of the optimal KL-constrained policy's advantage function. Using our notation for the relabeled policy and goal-conditioned setting, the AWR objective is:
\begin{equation*}
    \max_\pi \E_{\pi_r} \left[e^{( \mathcal{R}_{s,a,g}^{\pi_r} - V^{\pi_r}(s,g)} \log \pi(a|s,g)\right]
\end{equation*}
where $\mathcal{R}_{s,a}^{\pi_r}$ is the Monte-Carlo estimate of the empirical returns of $\pi_r$ taking action $a$ from state $s$. In practice, AWR uses bootstrapping to compute the advantage estimate as $r(s,a,g) + V^{\pi_r}(s',g) - V^{\pi_r}(s,g)$ instead of pure supervised learning. This corresponds to solving one step of the policy improvement objective from \cite{schulman2015trust}:
\begin{equation*}
    \eta(\pi) = \E_{\pi_r} \left[\mathcal{R}_{s,a,g}^{\pi_r} - V^{\pi_r}(s,g)\right] - \beta \E_{s\sim \pi_r} [D_{KL}(\pi || \pi_r)]
\end{equation*}
Note that with \abv we can reconstruct estimates of $V^{\pi_r}(s,g)$ by computing the expectation of the discrete distance distribution as follows:
\begin{equation*}
    V^{\pi_r}(s,g) = \E_{k\sim p^r(\cdot|s,g)} \left[\mathcal{R}_k \right]
\end{equation*}
We then recover a version of AWR using classification which performs a single-step of policy improvement.





\section{\abv Algorithm}
\label{app:alg}
Below we provide a detailed outline of the \abv algorithm.
\begin{algorithm}[H]
   \caption{\fullname}
   \label{alg}
\begin{algorithmic}
    \STATE {\bfseries Input:} Dataset $\mathcal{D}$, Goal function $\phi$, N-Step $N$, Horizon $T$, Temperatures $\alpha, \beta$
    \STATE {\bfseries Initialize:} Distribution $p^r_\theta$ with $B = T // N$ bins, Policy $\pi_\psi$
    \FOR{distribution training steps...}
    \STATE Sample relabeled data $(s_i, \phi(s_j)), j > i$
    \STATE Update $\theta$ via $\max_\theta \E_{\mathcal{D}_r}\left[\log p^r_\theta\left((j - i - 1)//N| s_i, \phi(s_j)\right)\right]$
    \ENDFOR
    \FOR{policy training steps...}
    \STATE Sample relabeled data $(s_i, s_{i+1}, \phi(s_j)), j > i$
    \STATE Compute $c(s_i, \phi(s_j)) = \indicator{\phi(c_{i+1}) \ne \phi(s_j)} / B$
    \STATE Compute $\hat{d}(s_i,\phi(s_j)) = -\alpha \log \E_{k \sim p^r_\theta(\cdot|(s_i,\phi(s_j))}\left[e^{-k/(B\alpha)}\right]$
    \STATE Compute $\hat{d}(s_{i+1},\phi(s_j)) = -\alpha \log \E_{k \sim p^r_\theta(\cdot|(s_{i+1},\phi(s_j))}\left[e^{-k/(B\alpha)}\right]$
    \STATE Compute $\text{adv} = \hat{d}(s_i,\phi(s_j)) - c(s_i, \phi(s_j)) - \hat{d}(s_{i+1},\phi(s_j))$
    \STATE Update $\psi$ via $\max_\psi \E_{\mathcal{D}_r}\left[e^{\text{adv}/\beta} \log \pi_\psi(a_i | s_i, \phi(s_j) )\right]$

    \ENDFOR
    \STATE {\bfseries Return:} $\theta, \psi$

\end{algorithmic}
\end{algorithm}

\section{Experiments}
\label{app:experiments}
We were unable to fit all experiments, ablations and learning curves within the body of the main paper. In this section, we include full learning curves for all experiments featured in the main paper, and ablations on more environments. This section is designed to mimic the main body of the paper.

\subsection{How does \abv perform?}

We primarily evaluated the performance of \abv on the following environments. In the main body of the paper, we omit state results for the Gym Robotics benchmarks from the main body of the paper for space. As expected, bootstrapping-based RL methods perform best on these benchmarks. 

\subsection*{Datasets and Environments}

\textbf{Gym Robotics}. In the Fetch environments, the agent is tasked with manipulating a block in different ways, namely pushing, pick and place, and sliding like in air hockey. The goal extraction function $\phi$ yields the location of the cube from the state vector. In the Hand Reach environment, a shadow hand robot is commanded to reach a specific hand configuration, and $\phi$ gives the locations of the fingers. For state-based experiments, we take the offline GCRL datasets from \citet{yang2022rethinking} which are either collected randomly, or from an expert policy with Gaussian noise of standard deviation 0.2. We use the same 90\% random, 10\% expert split from \citet{ma2022offline}. However, this type of dataset inherently disadvantages imitation learning methods, as one could expect to do better by just discarding the random data. Thus, we also create our own suboptimal datasets by training expert agents on each environment with TD3+HER, and add larger amounts of independent Gaussian noise, usually standard deviation 1 or 2. Note that the action space for these environments range from -1 to 1, thus this amount of noise results in very sub-optimal data. For the visual Gym Robotics environments, we also collect offline datasets using TD3+HER agents, but render the environment to RGB images of resolution 64x64. In all visual experiments $\phi$ is the identity. At test time we construct goal images by moving agents into a state deemed ``successful'' by the underlying state environment. Vision datasets are smaller, consisting of 250K transitions, except for Hand Reach where we used 500K transitions from a noisy expert, and 1 million transitions comprised of 90\% random interactions and 10\% expert. When constructing our own 90\% random, 10\% expert datasets, our expert policy also has Gaussian noise of standard deviation 0.2 as in \citet{yang2022rethinking}. In Table \ref{tab:dataset_stats} we provide return statistics of the Fetch datasets we created.

\begin{table}[]
\centering
\begin{tabular}{lcccc}
\multicolumn{1}{c}{\textbf{Dataset}}                           & \textbf{Mean} & \textbf{Mediam} & 7\textbf{5th \%ile} & \textbf{90th \%ile} \\ \hline
Fetch Push Image, 250K Noise 1.0  & 13.43                    & 8                          & 25                             & 35                             \\
Fetch Pick Image, 250K Noise 1.0  & 3.34                     & 0                          & 2                              & 11                             \\
Fetch Slide Image, 250K Noise 0.5 & 4.13                     & 0                          & 5                              & 15                             \\
Fetch Push Image, 250K Noise 2.0  & 3.28                     & 0                          & 1                              & 13                             \\
Fetch Pick Image, 250K Noise 2.0  & 0.7                      & 0                          & 0                              & 0                              \\
Fetch Slide Image, 250K Noise 1.0 & 1.73                     & 0                          & 1                              & 5                             
\end{tabular}
\caption{Here we show different return statistics of the noisy Fetch Image datasets that we create. As can be seen, the median trajectory return in 5 of 6 datasets is zero, showing that the datasets we test on are indeed extremely sub-optimal.}
\label{tab:dataset_stats}
\end{table}

\textbf{Franka Kitchen}. We take the Franka Kitchen demo dataset from \citet{gupta2020relay} and make it goal-conditioned as done in \citet{cui2022play} by choosing $\phi$ to be the identity. We use the same train/test split as done in \citet{cui2022play}, and use the final states of demo trajectories as goal states.

\textbf{Robomimic}. The Robomimic datasets from \citet{robomimic2021} are intended for unconditional behavior cloning. We make them goal-conditioned by setting $\phi$ to be the position of the object of interest, either the can for bin moving, or the square for peg insertion. We also discard the relative positions of the object to the robot end effector as we found including these components in the state hurt performance in the goal-conditioned setting. This corresponds to the latter seven components of the ``object'' key in robomimic. We use the same datasets, and set the evaluation horizon to 500.

\subsection*{Baselines}

We compare our method to the following baselines on all environments:

\textbf{GCSL.} This purely supervised technique from \citet{ghosh2021learning} uses hindisght relabeling with imitation learning losses. 

\textbf{WGCSL.} This method from \citet{yang2022rethinking} modifies GCSL by computing advantage weights using $Q$-Learning. WGCSL contains three weighting components: DRW which weights by the discount factor, GAEW which weights by the advantage from the learned Q function (this is the most important component), and BAW which retains only the best advantages over time. Following \citet{ma2022offline}, we implement WGCSL with BAW and GAEW, but fix an error in the advantage calculation used in \citet{ma2022offline}. This leads to WGCSL performing better than reported in \citet{ma2022offline}.

\textbf{GoFAR.} \citet{ma2022offline} propose using a state-matching objective for GCRL. They learn a reward discriminator and a bootstrapped value function for weighting the imitation learning loss. We found that this method was unstable on image datasets when using only actor gradients, causing some runs to crash. When reporting results we considered only up until the first seed crashed. This still resulted in better results than using gradients from the value function.

\textbf{GCIQL.} We modify IQL from \citet{kostrikov2022offline} to be goal-conditioned. IQL is a state-of-the-art offline RL algorithm that uses expectile regression to implicitly estimate maximal values. Unlike all other methods, GCIQL uses an actor, critic, and value network.

\textbf{DWSL-B.}
Instead of using supervised learning as in \abv, in this variant we learn distance distributions using bootstrapping, similar to the distributional $Q$-learning approach used in \citet{eysenbach2019search}. However, unlike in \citet{eysenbach2019search}, we only learn the behavioral distance distribution of our dataset, rather than perform $Q$-learning. Thus, to perform additional policy improvement, we still extract distance estimates using the LogSumExp as in \abv.

Below we include results for state-based Gym Robotics experiments in Table \ref{tab:state_results}, and full learning curves for all other experiments on different datasets.

\begin{figure}[H]
\includegraphics[]{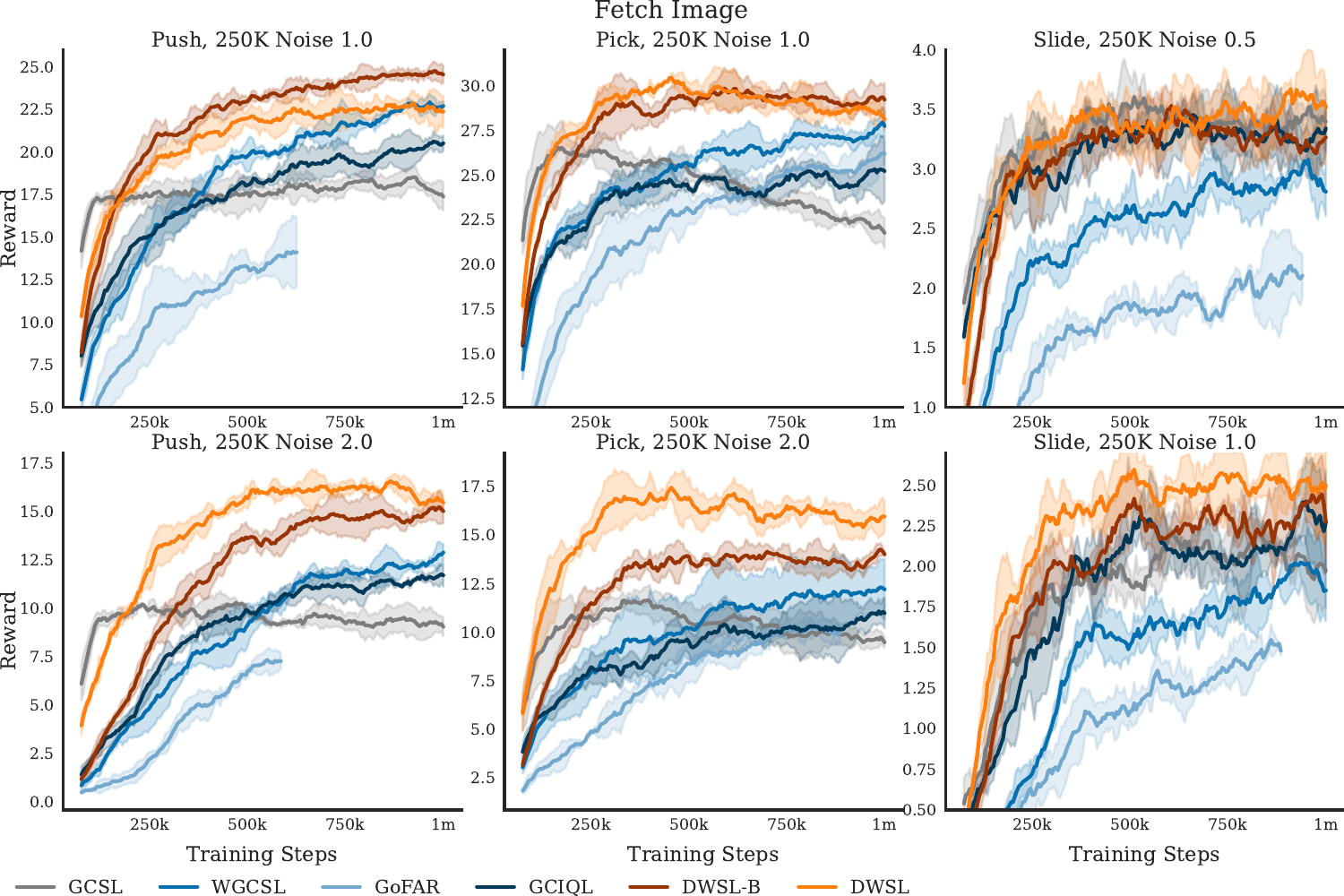}
\centering
\vspace{-0.1in}
\caption{Learning curves for the Fetch image datasets.}
\end{figure}

\begin{figure}[H]
\includegraphics[width=\textwidth]{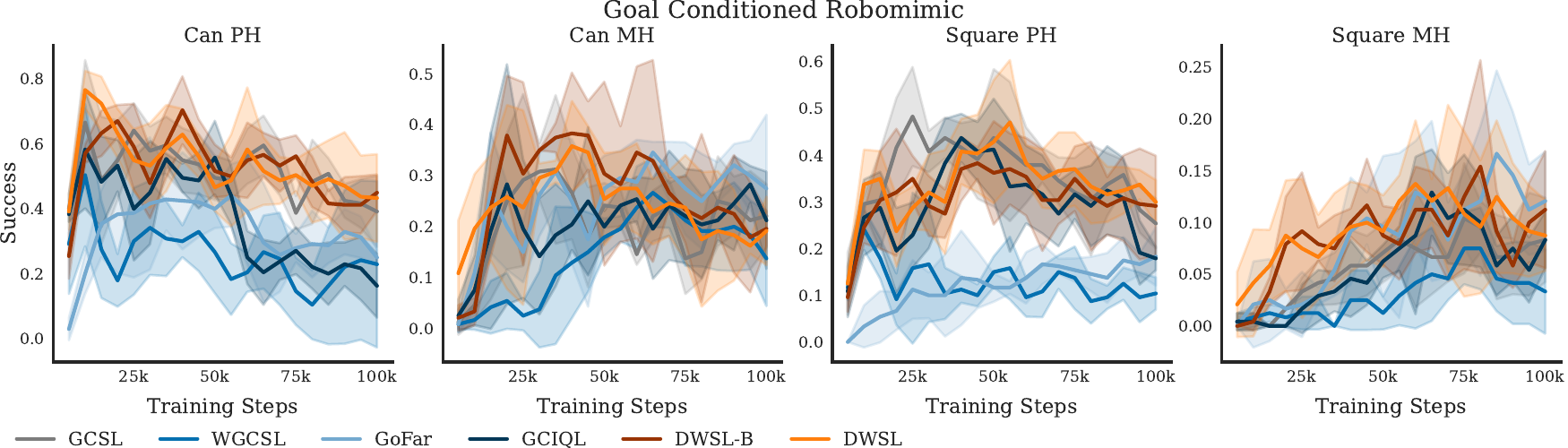}
\centering
\vspace{-0.2in}
\caption{Learning curves for Robomimic.}
\end{figure}

\begin{table*}
\resizebox{\textwidth}{!}{
\centering
\begin{tabular}{cccccccc}
\textbf{Environment}               & \textbf{Dataset} & \textbf{GCSL}           & \textbf{WGCSL}  & \textbf{GoFAR}           & \textbf{GCIQL}  & \textbf{DWSL-B}         & \textbf{DWSL}           \\ \hline
\multirow{2}{*}{\shortstack{\textbf{Fetch Push} \\ \includegraphics[width=0.5in,page=1]{figures/icons.pdf}}}  & \multicolumn{1}{l|}{2M 90\% R 10\% E}     & 27.97 $\pm$ 1.17          & \textbf{37.14 $\pm$ 0.68}  & \textbf{37.56 $\pm$ 0.50}          & \textbf{36.65 $\pm$ 0.59}  & 34.57 $\pm$ 0.96 & 31.60 $\pm$ 0.99          \\
                                   & \multicolumn{1}{l|}{500K Noise 2}     & 17.88 $\pm$ 0.29          & \textbf{29.87 $\pm$ 1.56}  & 27.85 $\pm$ 1.41            & \textbf{29.67 $\pm$ 1.12}  & 25.51 $\pm$ 1.80          & 24.23 $\pm$ 1.63 \\[0.2cm] 
\multirow{2}{*}{\shortstack{\textbf{Fetch Pick} \\ \includegraphics[width=0.5in,page=2]{figures/icons.pdf}}}  & \multicolumn{1}{l|}{2M 90\% R 10\% E}     & 23.81 $\pm$ 1.00         & \textbf{34.90 $\pm$ 1.03}  & \textbf{33.36 $\pm$ 1.63}           & \textbf{34.23 $\pm$ 1.61} & 32.37 $\pm$ 0.91 & 30.51 $\pm$ 1.61 \\
                                   & \multicolumn{1}{l|}{500K Noise 2}     & 11.37 $\pm$ 0.98          & \textbf{19.76 $\pm$ 3.57} & 17.03 $\pm$ 2.09          & \textbf{19.37 $\pm$ 0.70}  & 18.50 $\pm$ 1.68          & 17.44 $\pm$ 2.12 \\[0.2cm]
\multirow{2}{*}{\shortstack{\textbf{Fetch Slide} \\ \includegraphics[width=0.5in,page=3]{figures/icons.pdf}}} & \multicolumn{1}{l|}{2M 90\% R 10\% E}   & 4.73 $\pm$ 0.56  & \textbf{9.50 $\pm$ 0.48}   & 6.82 $\pm$ 0.94            & \textbf{9.13 $\pm$ 0.49}   & 8.14 $\pm$ 0.58  & 8.06 $\pm$ 0.68  \\
                                   & \multicolumn{1}{l|}{500K Noise 0.2}     & 3.95 $\pm$ 0.40           & \textbf{11.85 $\pm$ 0.50}   & 7.88 $\pm$ 0.42            & 11.25 $\pm$ 0.77   & 9.90 $\pm$ 0.52           & 10.03 $\pm$ 0.33  \\[0.2cm]
\multirow{2}{*}{\shortstack{\textbf{2M 90\% R 10\% E} \\ \includegraphics[width=0.5in,page=4]{figures/icons.pdf}}}  & \multicolumn{1}{l|}{1M 90\% R 10\% E} & 6.39 $\pm$ 0.69          & \textbf{26.98 $\pm$ 2.51}  & 18.82 $\pm$ 3.68           & 17.55 $\pm$ 2.67  & 19.10 $\pm$ 0.80          & 18.13 $\pm$ 2.17 \\
                                   & \multicolumn{1}{l|}{500K Noise 0.2}   & 0.95 $\pm$ 0.54          & \textbf{30.30 $\pm$ 0.55}  & 28.57 $\pm$ 2.47 & \textbf{30.08 $\pm$ 1.14}  & 27.34 $\pm$ 1.56        & 26.41 $\pm$ 1.31       
\end{tabular}
}
\vspace{-0.15in}
\caption{Results for state-based Gym Robotics datasets. Bolded numbers are within 95\% of the best score. We run four seeds in state-based domains. While bootstrapping-based RL methods perform best on these datasets, \abv still exhibits policy improvement and consistently outperforms GCSL.}
\label{tab:state_results}
\vspace{-0.1in}
\end{table*}

\begin{minipage}[b]{0.35\linewidth}
\begin{figure}[H]
\centering
\includegraphics[width=\textwidth]{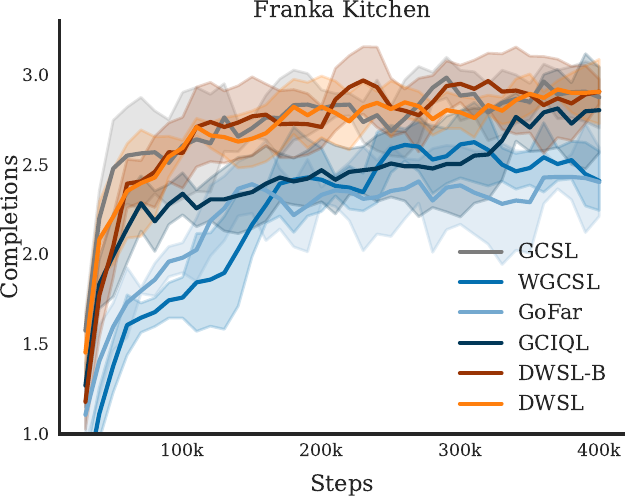}
\vspace{-0.28in}
\caption{Learning curve for Franka Kitchen}
\end{figure}
\end{minipage}
\hspace{0.5cm}
\begin{minipage}[b]{0.61\linewidth}
\begin{figure}[H]
\centering
\includegraphics[width=\textwidth]{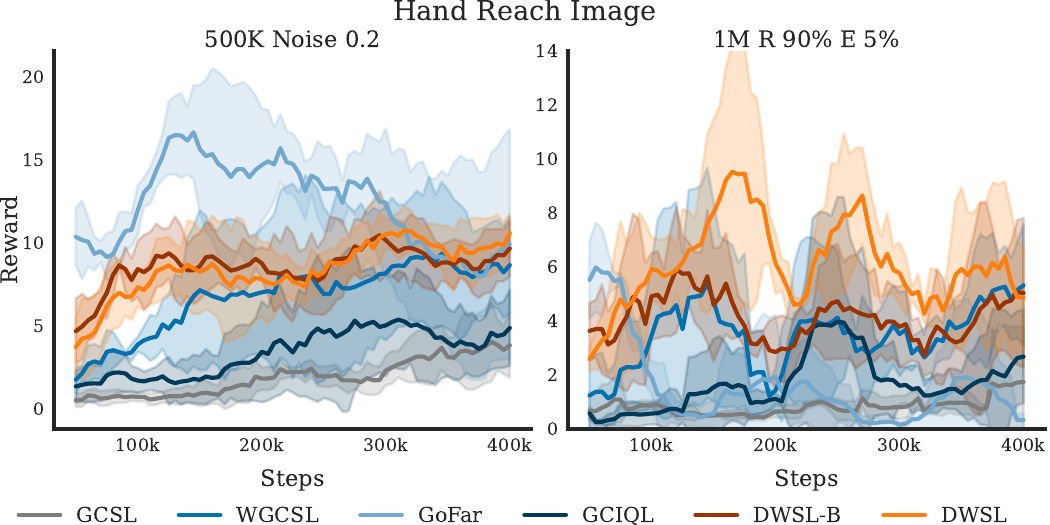}
\vspace{-0.28in}
\caption{Learning curves for the Hand Reach image datasets.}
\end{figure}
\end{minipage}



\subsection{Does \abv use the right objective?}
\begin{figure}[H]
\centering
\includegraphics[width=0.69\textwidth]{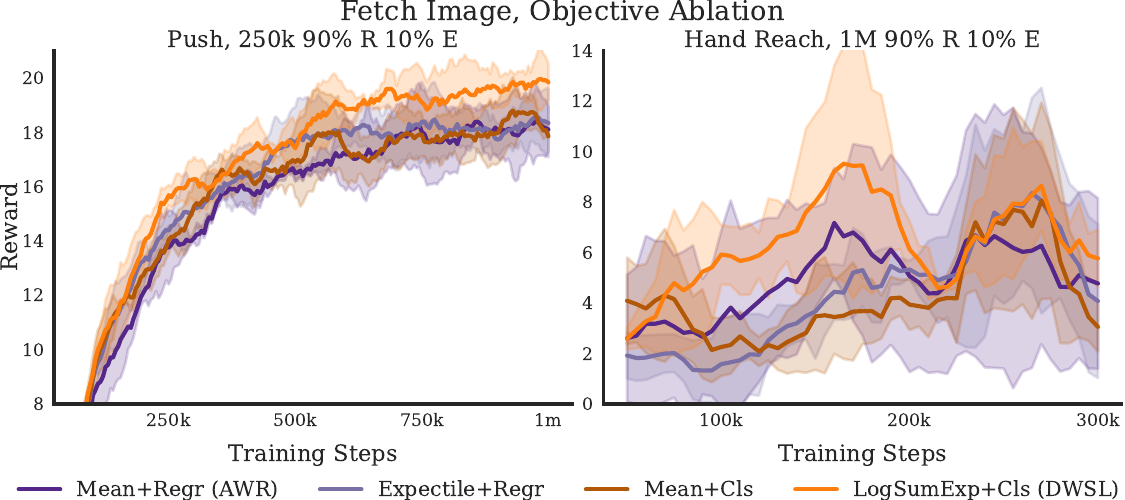}
\vspace{-0.1in}
\caption{Objective ablation on image datasets.}
\label{fig:objective_pixels}
\end{figure}

\begin{figure}[H]
\includegraphics[width=0.69\textwidth]{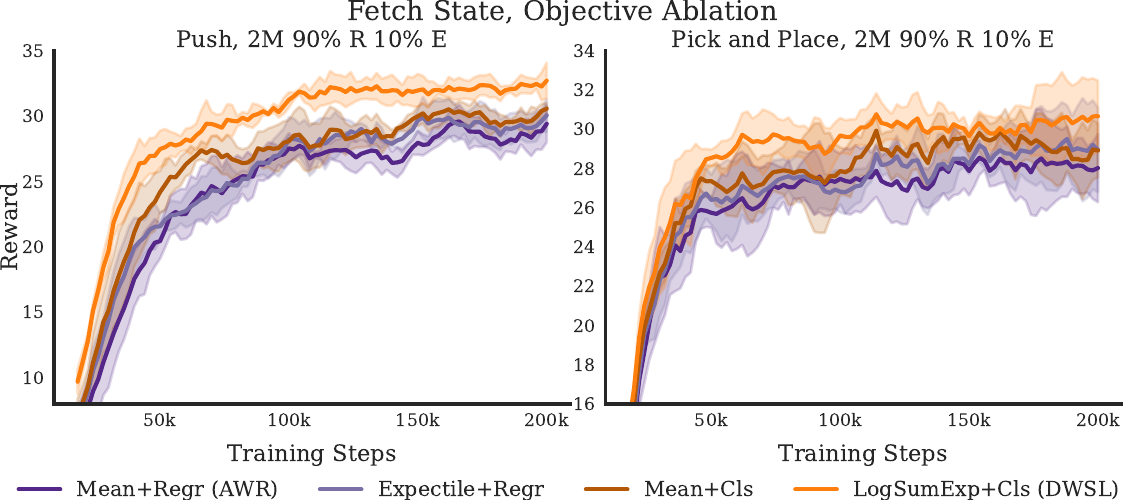}
\centering
\caption{Objective ablation on state datasets.}
\label{fig:objective_state}
\end{figure}

In this section we compare \abv's supervised objective to different possibilities. We find the most impact on datasets with more biased actions, i.e. where the mean action for each state is less optimal. For datasets comprised of noisy rollouts of a trained policy, where actions are less biased, we suspect that the average computed distance is highly correlated with the optimal distance. When we consider datasets with random data as well, which introduces more action bias, the advantage of \abv's objective becomes more clear. Specifically, these datasets contain $90\%$ Random data, and $10\%$ Expert data with 0.2 standard deviation Gaussian noise. For the state based dataset with contain $90\%$ Random data, and $10\%$ Expert data, we decrease the value of $\alpha$ from $1$ to $0.1$. This lets \abv be less constrained to the random data. Like in our main paper, the image environments stay at $\alpha =1$. Results can be found in Figures \ref{fig:objective_state} and \ref{fig:objective_pixels}. In the next section, we also include results for different supervised learning methods on AntMaze.

\subsection{When does \abv fail?}
In this section we consdier the AntMaze task from \citet{fu2020d4rl}. In AntMaze, the agent learn to navigate mazes of varying sizes by stitching together offline trajectories. While this task is usually learned from rewards, we make it goal-conditioned by setting $\phi$ to be the $(x, y)$ position of the agent in the maze. Below, we include full learning curves for all methods on AntMaze. We also include DWSL-B, which was omitted from the main paper for space. In Table \ref{tab:antmaze_objective}, we also include results for different supervised objectives on the AntMaze benchmark. These results, which show that all supervised methods suffer on the AntMaze stitching data. This indicates that our use of classification over regression is not inhibiting \abv's performance in these settings.

\begin{figure}[H]
\includegraphics[]{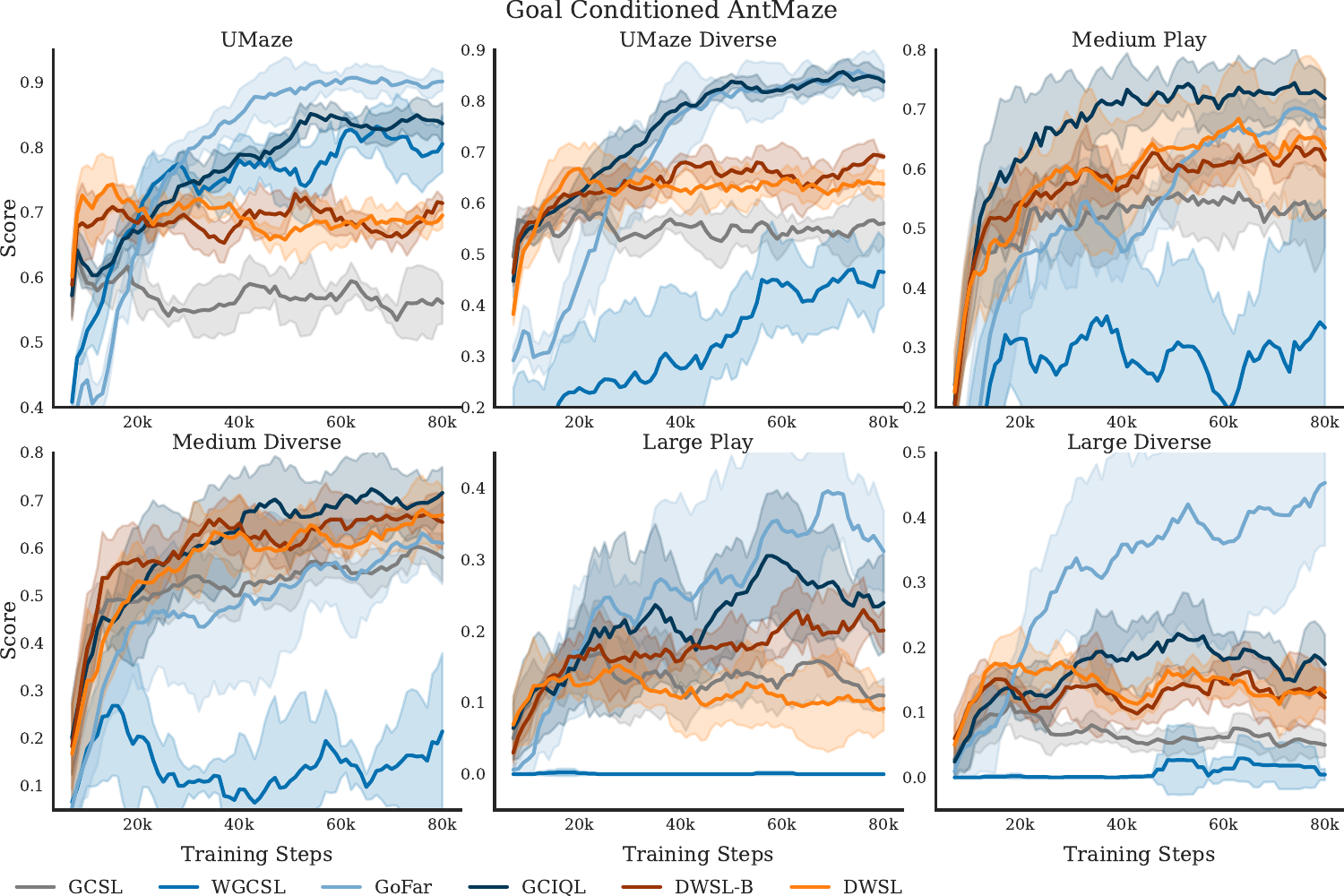}
\centering
\caption{Learning curves for the AntMaze datasets.}
\end{figure}

\begin{table}[h]
\centering
\begin{tabular}{lcccc}
\textbf{Dataset}        & \textbf{GCSL}       & \textbf{AWR} & \textbf{Expectile+Regr} & \textbf{DWSL} \\ \hline
UMaze          & 64 $\pm$ 2 & 68 $\pm$ 4      & 72 $\pm$ 2     & \textbf{74 $\pm$ 4}             \\
UMaze Diverse  & 59 $\pm$ 1 & 61 $\pm$ 5      & 60 $\pm$ 6     & \textbf{67 $\pm$ 3  }           \\
Medium Play    & 56 $\pm$ 6 & 59 $\pm$ 7      & 61 $\pm$ 4     & \textbf{69 $\pm$ 8 }            \\
Medium Diverse & 60 $\pm$ 3 & 61 $\pm$ 4      & 59 $\pm$ 12    & \textbf{68 $\pm$ 6  }           \\
Large Play     & \textbf{17 $\pm$ 5} & 15 $\pm$ 3      & \textbf{17 $\pm$ 7}     & 15 $\pm$ 5             \\
Large Diverse  & 12 $\pm$ 3 & 15 $\pm$ 6      & 17 $\pm$ 4     & \textbf{18 $\pm$ 2}            
\end{tabular}
\caption{Performance on the AntMaze benchmark with different supervised learning algorithms. \abv maintains the highest performance overall.}
\label{tab:antmaze_objective}
\end{table}

\subsection{How robust is \abv?}
\label{app:ablations}
In this section we evaluate the robustness of \abv to various properties, such as architecture size, hyperparameters, and relabeling ratio. Due to space limitations, we did not include all of these ablations in the main body of the paper. We include them below. 

\begin{figure}[h]
\includegraphics[]{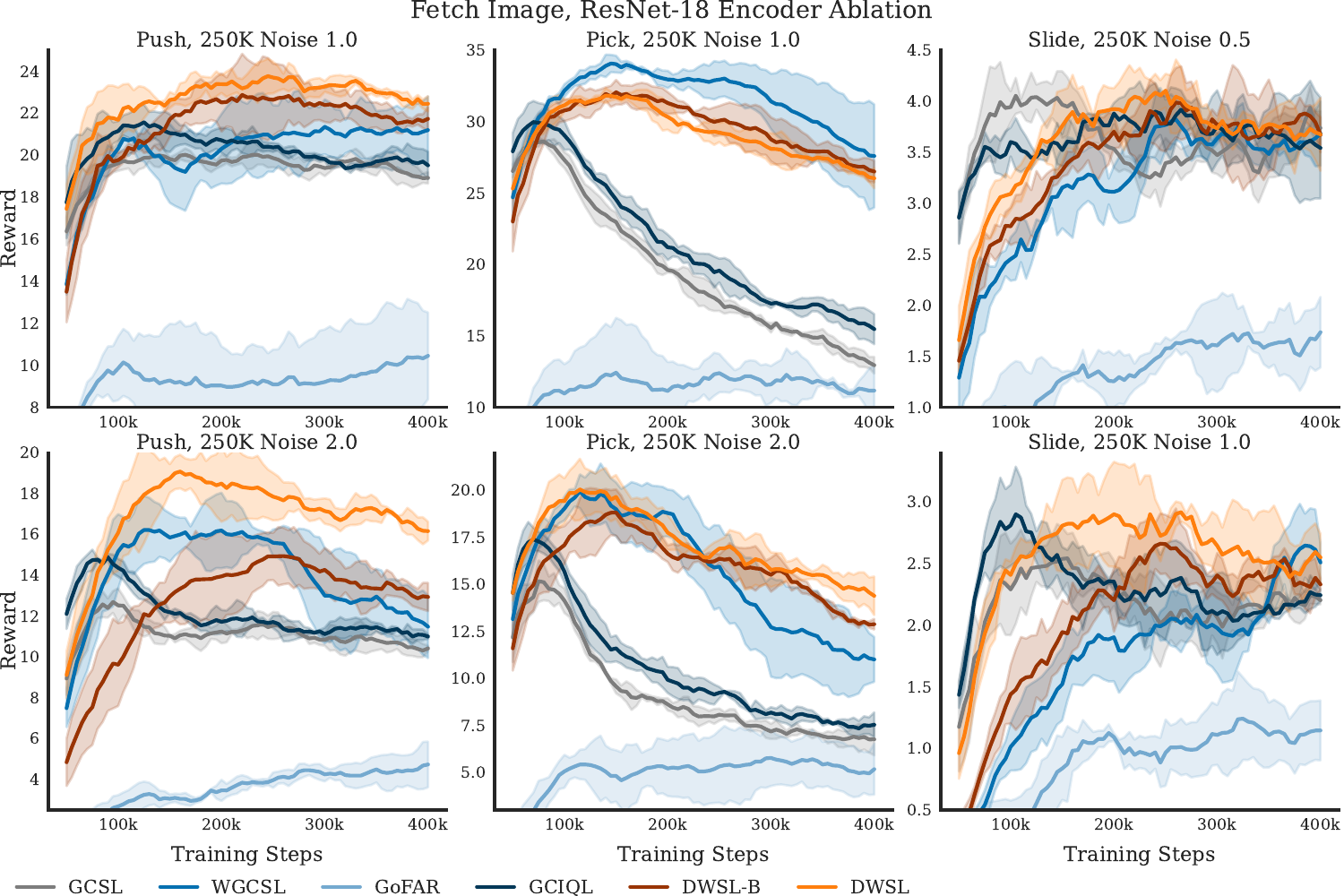}
\centering
\caption{Learning curves for the Fetch image datasets using the ResNet18 Architecture from RoboMimic.}
\label{fig:fetch_resnet}
\end{figure}

\textbf{Larger Architectures.} We experiment with using larger ResNet18-based encoders with the Robomimic architecture \citep{robomimic2021}. We concatenate current and goal images along the channel axis. Overall, we find improved performance and faster convergence (or overfitting) with larger architectures. Full results with ResNet architectures on the Gym Fetch Image datasets are shown in Figure \ref{fig:fetch_resnet}.

\textbf{Image Gradients.} We tried training our image encoders using gradients from both the actor and value/distance function for different offline GCRL algorithms. Interestingly, we got better performance when using only gradients from the actor network. We believe that this is because supervised learning and action prediction provide a stronger representation learning signal. We present results ablating this decision on the ``Fetch Push Noise 2.0'' dataset below, where actor-only gradients outperform or match ``both'' gradients for all methods. We therefore use actor-only gradients on all methods in image domains for consistency.

\begin{table}[H]
\centering
\begin{tabular}{lll}
\multicolumn{1}{c}{\textbf{Algorithm}}          & \multicolumn{1}{c}{\textbf{Actor}}   & \multicolumn{1}{c}{\textbf{Both}}    \\ \hline
\multicolumn{1}{l|}{WGCSL}  & \textbf{12.89} $\pm$ 0.36 & 11.27 $\pm$ 1.68 \\
\multicolumn{1}{l|}{GoFAR}  & \textbf{7.34 $\pm$ 0.44}  & 5.71 $\pm$ 0.93  \\
\multicolumn{1}{l|}{GCIQL}  & \textbf{11.76 $\pm$ 0.38} & \textbf{11.76 $\pm$ 0.48} \\
\multicolumn{1}{l|}{DWSL-B} & \textbf{15.23 $\pm$ 0.70} & \textbf{15.51 $\pm$ 0.47} \\
\multicolumn{1}{l|}{DWSL}   & \textbf{16.57 $\pm$ 0.18} & \textbf{16.34 $\pm$ 0.02}
\end{tabular}
\caption{Comparing learning the image encoder using actor-only gradients, or ``both'' gradients (where a learned value/distance function also provides gradients), on the ``Fetch Push Noise 2.0'' dataset.}
\end{table}

\begin{figure}[]
\includegraphics[]{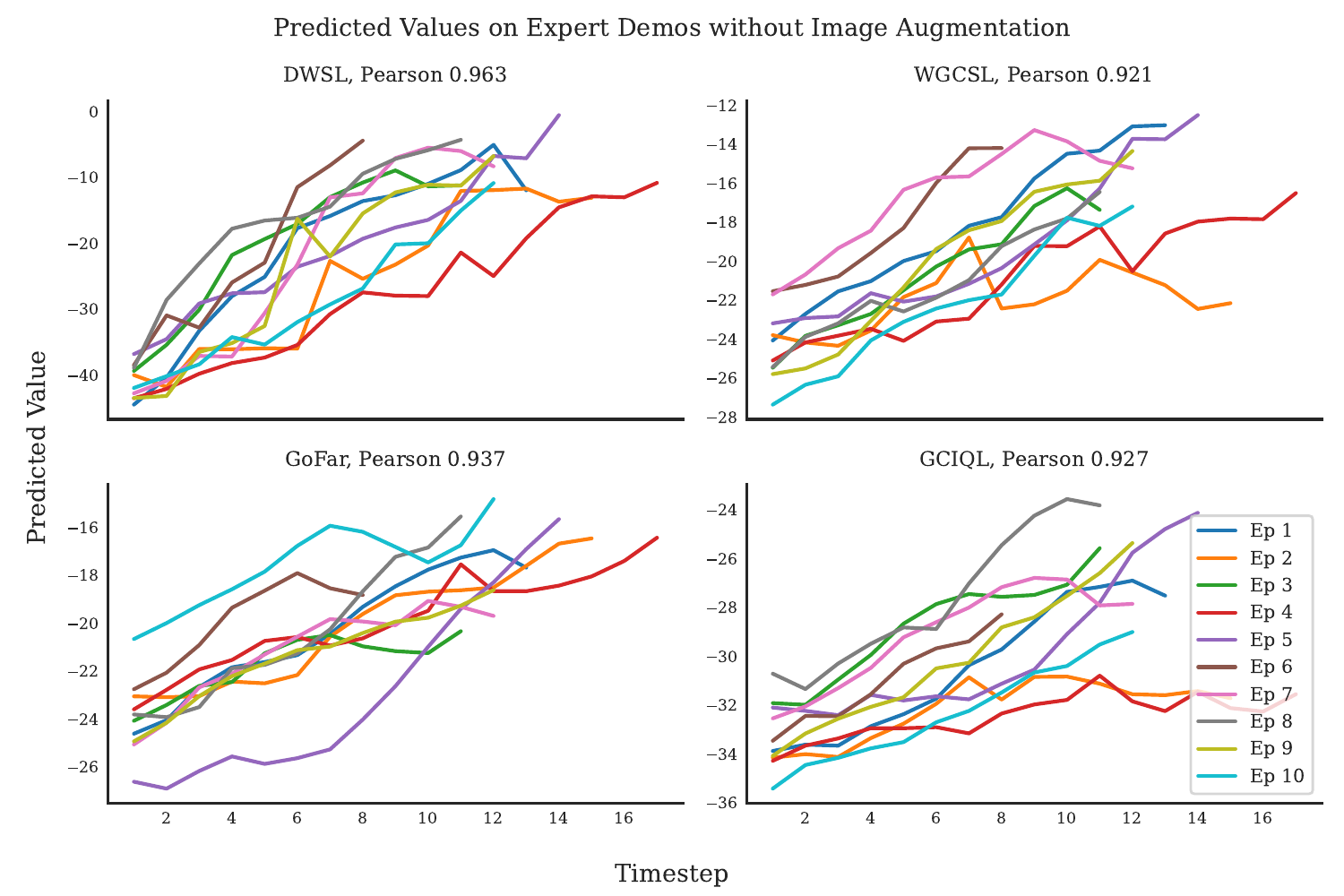}
\centering
\caption{Plots of learned values versus timesteps on ten expert trajectories without any image augmentation.}
\label{fig:push_corr_no_aug}
\end{figure}

\begin{figure}[]
\includegraphics[]{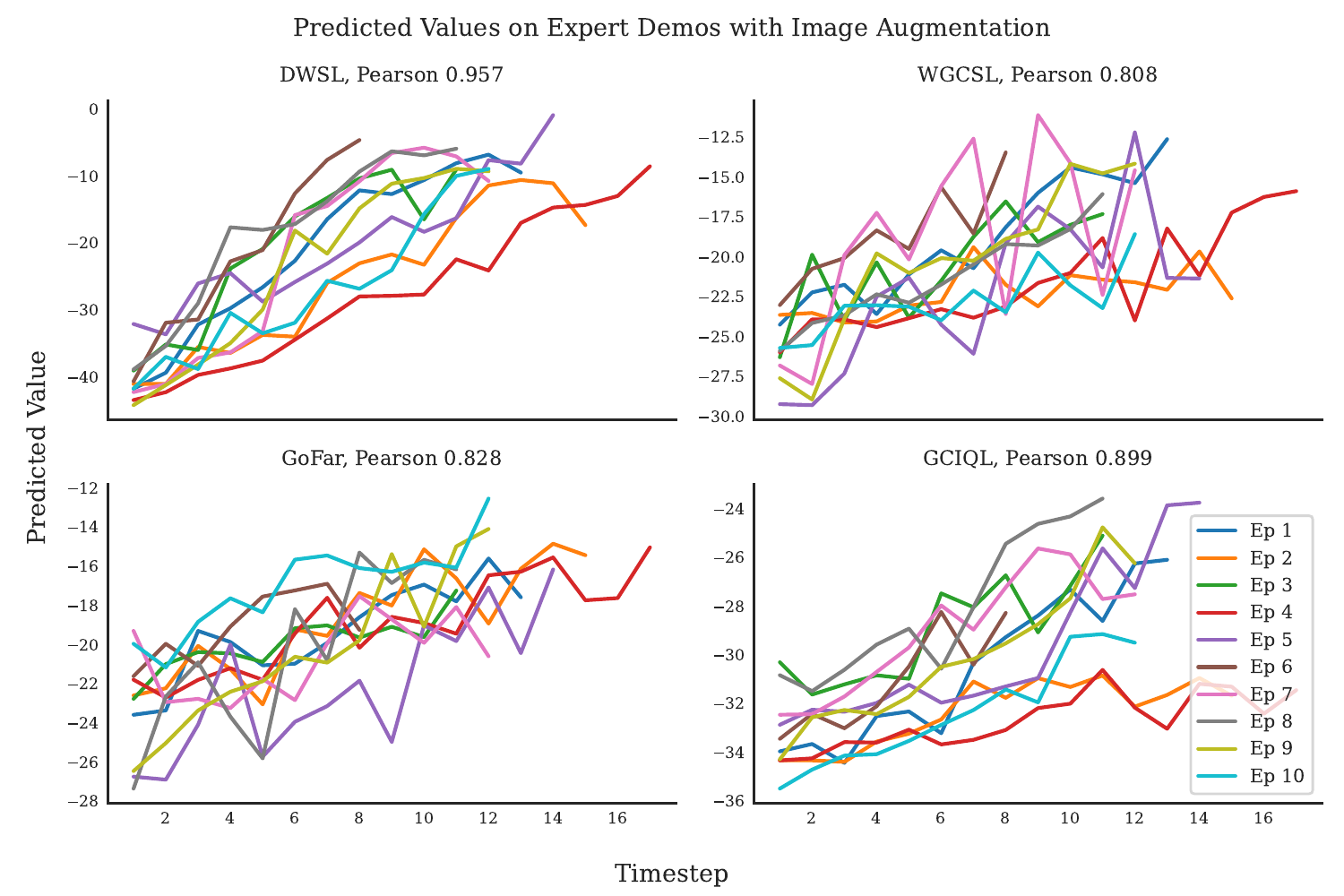}
\centering
\caption{Plots of learned values versus timesteps on ten expert trajectories with random image cropping augmentation.}
\label{fig:push_corr_aug}
\end{figure}

\textbf{Robustness to Image Augmentations.} Value functions learned by \abv exhibit more robustness to random image shifting and cropping augmentations than those learned by temporal difference based methods. We test this by taking ten expert trajectories on the Fetch Push Image task and plotting the value function learned from the ``Fetch Push Noise 2.0'' dataset at each state sequentially. A perfect value function would almost linearly increase from the start of the trajectory to the end. To quantitatively measure this, we compute the Pearson correlation coefficient of predicted values with the timesteps in each demonstration. Our results are summarized in Table \ref{tab:aug_robustness} and depicted in Figures \ref{fig:push_corr_no_aug} and \ref{fig:push_corr_aug}. \abv has the highest correlation coefficient both with and without augmentations. When we add image augmentations, the correlation coefficient of temporal difference based methods drops significantly. Qualitatively, we see a much worse linear relationship. This again indicates the added robustness gained by using supervised learning instead of bootstrapping. 

\begin{table}[h]
\centering
\begin{tabular}{lcccc}
            & \textbf{WGCSL} & \textbf{GoFar} & \textbf{GCIQL} & \textbf{DWSL} \\ \hline
No Aug      & 0.921                     & 0.937                     & 0.927                     & \textbf{0.963}           \\
Random Crop & 0.808                     & 0.828                     & 0.899                     & \textbf{0.957}           \\
Diference   & 0.113                     & 0.109                     & 0.028                     & \textbf{0.006}          
\end{tabular}
\caption{Correlation coefficients with and without image augmentations for ten expert trajectories on the Fetch Push Image Noise 2.0 dataset.}
\label{tab:aug_robustness}
\end{table}

\textbf{Robustness to Hyperparameters.} Because \abv uses only supervised learning, we posit that it will exhibit a higher level of robustness to hyperparameters than $Q$-Learning approaches. We try $\alpha \in [0.1, 1, 10]$ and $\beta = [0.01, 0.05, 0.25]$ in state-based Gym Robotics environemnts. Different values of $\alpha$ tradeoff how conservative \abv is. Thus, for some datasets lower values of $\alpha$ may work better, particularly when there is more random data. We include full results on all tasks with $\alpha = 1$ in Table \ref{tab:alpha}.

\begin{figure}[H]
\includegraphics[]{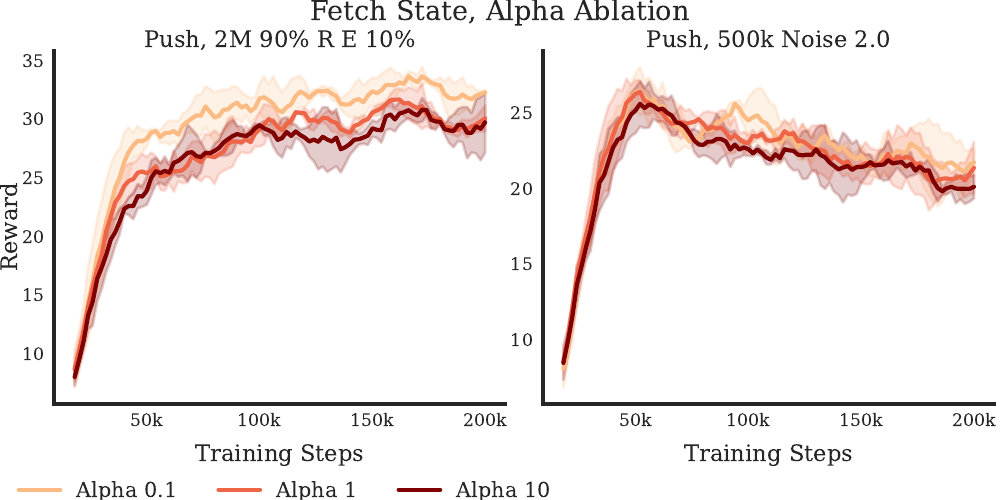}
\centering
\vspace{-0.1in}
\caption{$\alpha$ ablation on Fetch Push state datasets.}
\end{figure}

\begin{figure}[H]
\includegraphics[]{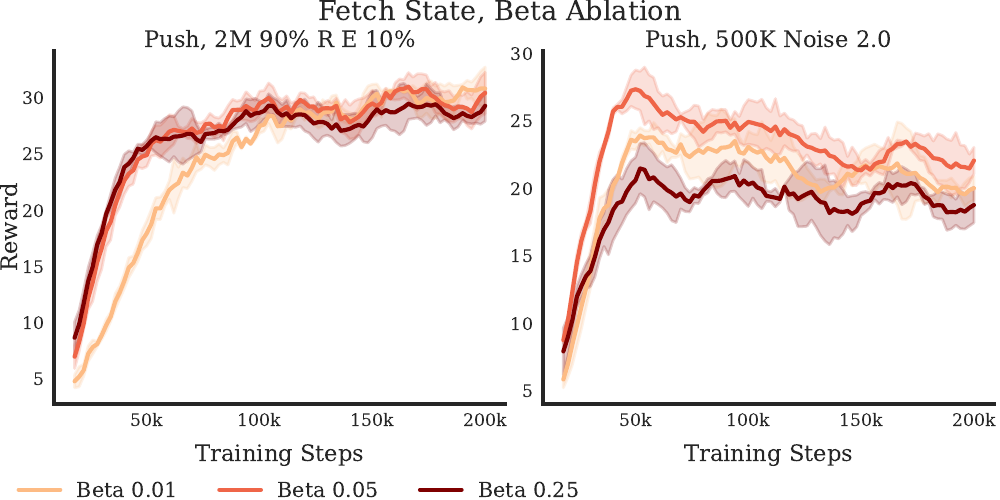}
\centering
\vspace{-0.1in}
\caption{$\beta$ ablation on Fetch Push state datasets.}
\end{figure}

\begin{table}[]
\centering
\begin{tabular}{lcc}
\multicolumn{1}{c}{\textbf{Dataset}} & \multicolumn{1}{c}{\textbf{$\alpha = 1$}} & \multicolumn{1}{c}{\textbf{$\alpha = 0.1$}} \\ \hline
Fetch Push Image, 250K Noise 1.0     & 22.88 $\pm$ 0.61                     & 24.07 $\pm$ 0.33                       \\
Fetch Pick Image, 250K Noise 1.0     & 30.46 $\pm$ 0.23                     & 29.53 $\pm$ 0.16                       \\
Fetch Slide Image, 250K Noise 0.5    & 3.67 $\pm$ 0.07                      & 3.61 $\pm$ 0.16                        \\
Fetch Push Image, 250K Noise 2.0     & 16.57 $\pm$ 0.18                     & 14.70 $\pm$ 0.23                       \\
Fetch Pick Image, 250K Noise 2.0     & 17.45 $\pm$ 0.47                     & 15.84 $\pm$ 0.59                       \\
Fetch Slide Image, 250K Noise 1.0    & 2.60 $\pm$ 0.13                      & 2.72 $\pm$ 0.12                        \\
Hand Reach Image, 500K Noise 0.2     & 10.79 $\pm$ 1.20                     & 10.20 $\pm$ 2.74                       \\
Hand Reach Image, 1M 90\% 10\% E     & 9.53 $\pm$ 3.82                      & 6.57 $\pm$ 0.98                        \\
Franka Kitchen                       & 2.92 $\pm$ 0.04                      & 2.88 $\pm$ 0.10                        \\
RoboMimic Can, 100 PH                & 77 $\pm$ 5                           & 70 $\pm$ 11                            \\
RoboMimic Can, 300 MH                & 36 $\pm$ 8                           & 32 $\pm$ 7                             \\
RoboMimic Square, 100 PH             & 47 $\pm$ 12                          & 47 $\pm$ 5                             \\
RoboMimic Square, 300 MH             & 14 $\pm$ 5                           & 13 $\pm$ 5                             \\
Antmaze UMaze                        & 74 $\pm$ 4                           & 76 $\pm$ 4                            
\end{tabular}
\caption{Comparison between $\alpha = 1$ and $\alpha = 0.1$ for all the tasks we study}
\label{tab:alpha}
\end{table}

\textbf{Relabeling Ratio}. While the \textit{learning from offline interaction} setting we deal with assumes data does not come with goal labels, here we investigate the effect of having goal labels on performance. We find that \abv is relatively robust to the relabeling ratio. On the other hand, $Q$-learning methods like WGCSL seem to be more sensitve. We provide analysis of this in the learning curves below.

\begin{figure}[H]
\includegraphics[width=\textwidth]{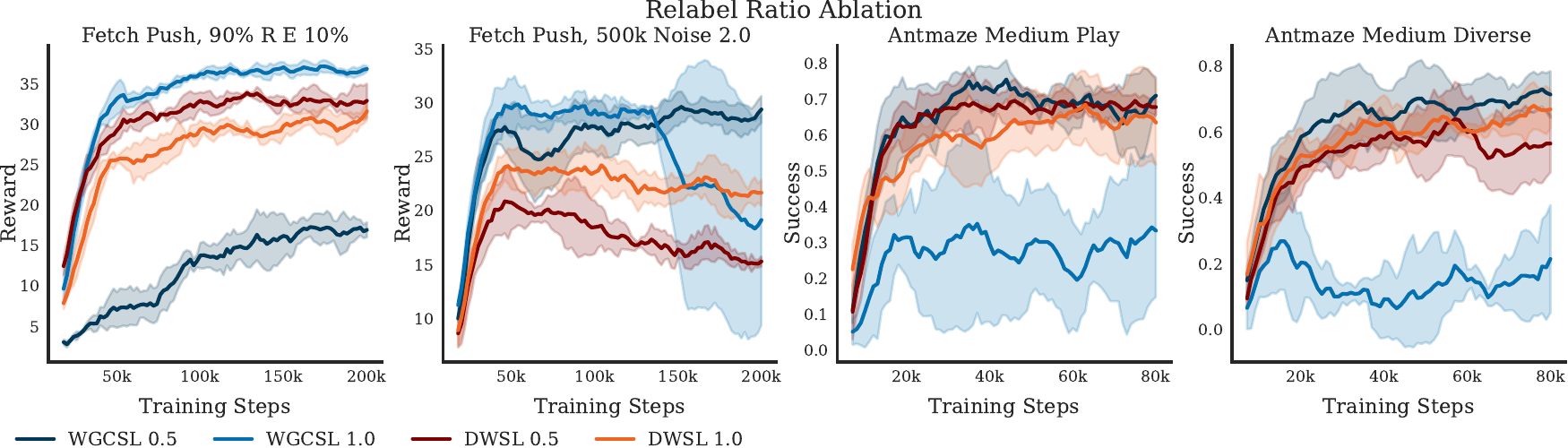}
\centering
\caption{Relabel ratio ablation on Fetch Push and AntMaze Medium Diverse and Play.}
\end{figure}

\begin{figure}[H]
\includegraphics[]{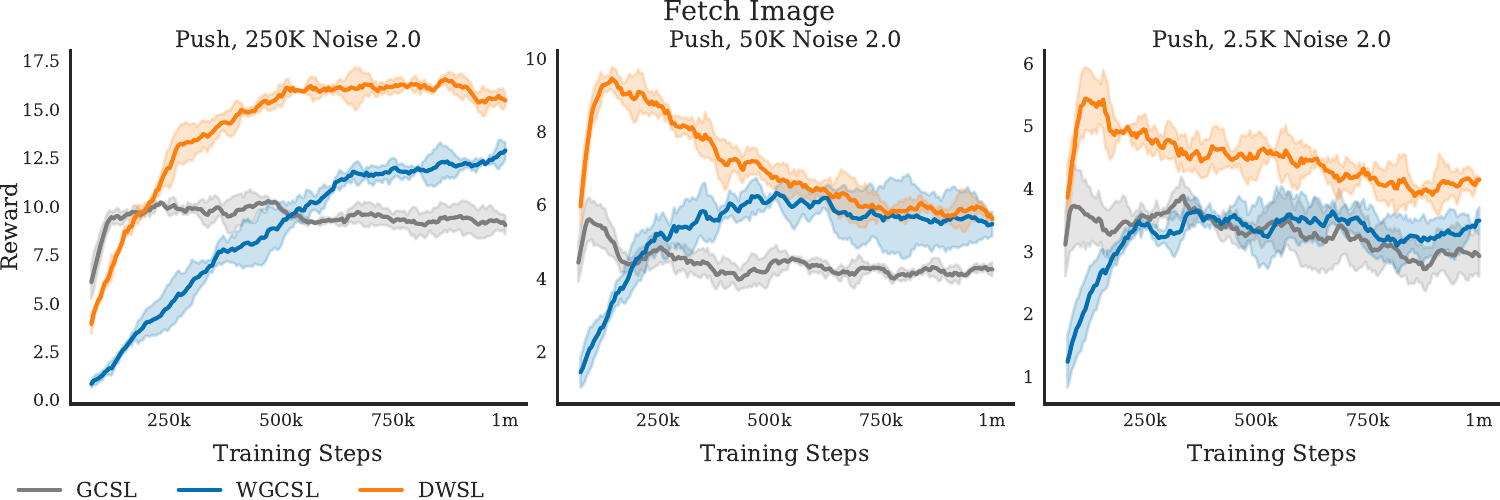}
\centering
\caption{Dataset size comparison on the Fetch Push task from Pixels.}
\label{fig:dataset_size}
\end{figure}

\textbf{Performance Versus Dataset Size}. We test how the performance of different methods vary when changing the size of the dataset for Visual Fetch Push with noise standard deviation 2.0 and include the results in Figure \ref{fig:dataset_size}. We find that the performance of all methods significantly drops as the size of the dataset is reduced. \abv maintains the highest performance at all dataset sizes.

\section{Implementation Details}
\label{sec:details}

\subsection{Architectures}

\textbf{State.} For all state-based experiments, we use MLPs with ReLU activations. For the policy network, we include a final tanh activation to normalize the outputs to each environment's action space of [-1, 1]. For all Gym Robotics environments, we use three hidden layers of size 256 as done in prior work \citep{yang2022rethinking,ma2022offline}. For all other state environments, we use two hidden layers of size 512, as larger networks have been shown to perform better for BC on Franka Kitchen, AntMaze, and Robomimic \citep{emmons2022rvs,robomimic2021}.

\textbf{Image.}
For our main image-based experiments, we adapt the network architectures used in \citet{yarats2021improving}. The main difference is because we operate in the goal-conditioned setting, the input to our convolutional encoder is the concatenation of the current image observation and goal image along the channel dimension. Consequently, we adjust the size of the initial convolutional filters to be compatible with a larger channel dimension. We share the visual encoder across all networks used for each algorithm. Representations from the encoder are first fed to a trunk module (specific to each network) consisting of a single fully-connected layer normalized by LayerNorm, and then tanh applied to the 50 dimensional output of this layer. All downstream networks consist of 2-layer MLPs with ReLU activations, with hidden dimension 1024. The policy networks again use tanh activations for output normalization.

For our ablation experiments on encoder architecture, we replace the previously described convolutional encoder with the ResNet-18 based encoder used in \citet{robomimic2021}, which consists of ResNet-18 followed by a spatial-softmax layer \citep{finn2016deep} with 64 keypoints. We similarily adjust the initial convolutional filters of the ResNet-18 to support the concatenated image observations, and the rest of its parameters are initialized from ImageNet pre-training. We remove the trunk module, but otherwise the other components of each network remain identical.

\subsection{Model Selection}
Model selection is a challenging problem in offline RL \citep{robomimic2021,emmons2022rvs}, particularly as validation loss does not always correlate with performance. As we compare supervised algorithms, like our method, with bootstrapping based methods, model selection is even more complicated. This is because supervised methods tend to train faster but can also overfit the data more easily, while RL approaches usually take longer to converge. Prior works in IL \citep{robomimic2021} compute the maximum evaluation performance of each seed, and average the maximums. This type of evaluation can be too optimistic about perfect model selection. Prior works in RL simply report average peformance after a fixed number of steps. If we trained RL methods to convergence, the performance of supervised methods, like \abv, could deteriorate from overfitting, giving an unfair advantage to RL approaches. To balance both of these interests, we make learning curves for all methods by averaging results across multiple seeds. We then report the mean and standard deviation of the highest point on the curves. This is equivalent to early-stopping for each method at the point where it performs best. Because of this, the evaluation frequency and number of evaluation episodes can influence the results. We keep this the same across all methods per dataset, and report the values we used for each environment in Table \ref{tab:domain_hparams}.

\subsection{Hyperparameters}
We use the Adam optimizer for all experiments. For all methods, we relabel goals for each state by sampling uniformly from all future states in its trajectory. For baselines that use discount factors, we use $\gamma = 0.98$ for Gym Robotics environments, and $\gamma = 0.99$ for the remaining environments. We ran four seeds for all state experiments, and 3 seeds for all image experiments. For algorithm specific hyperparameters, we include them in Table \ref{tab:alg_hparams}. ``Max clip'' refers to the maximum exponentiated advantage weight we clip to. For other hyperparamters common to all methods, we include them in Table \ref{tab:domain_hparams}.

For DWSL-B and \abv, we tuned $\beta$ using state-based Fetch Push. We also tuned the expectile value of GCIQL using state-based Fetch Push. We did not tune $\alpha$ for DWSL-B or \abv. Other algorithm specific hyperparameters are default values used in past work \citep{yang2022rethinking, ma2022offline}.

\begin{table}[]
\centering
\begin{tabular}{l|lr}
\textbf{Algorithm} & \textbf{Hyperparameter} & \multicolumn{1}{l}{\textbf{Value}}       \\ \hline
WGCSL              & $\beta$                 & 1                                        \\
                   & Max clip                & 10                                       \\
                   & Target update frequency & 20                                       \\
                   & Polyak coefficient      & \multicolumn{1}{l}{0.05 (0.1 on images)} \\ \hline
GoFAR              & $\beta$                 & 1                                        \\
                   & Max clip                & 10                                       \\
                   & Target update frequency & 20                                       \\
                   & Polyak coefficient      & 0.05                                     \\ \hline
GCIQL              & $\beta$                 & 1                                        \\
                   & Max clip                & 10                                       \\
                   & Target update frequency & 20                                       \\
                   & Polyak coefficient      & 0.05                                     \\
                   & Expectile               & 0.7                                      \\ \hline
DWSL-B             & $\beta$                 & 0.05                                     \\
                   & Max clip                & 10                                       \\
                   & Target update frequency & 20                                       \\
                   & Polyak coefficient      & 0.05                                     \\
                   & $\alpha$                & 1                                        \\ \hline
DWSL               & $\beta$                 & 0.05                                     \\
                   & Max clip                & 10                                       \\
                   & $\alpha$                & 1                                       
\end{tabular}
\caption{Algorithm Specific Hyperparameters} 
\label{tab:alg_hparams}
\end{table}

\subsection{Rewards}
\textbf{State.}
For methods that use rewards (namely offline GCRL), we assume access to a sparse reward function $r(s, a, g) = -\indicator{||\phi(s) - g||_2 < d} $ that detects whether a state is within some distance threshold $d$ of a goal. These distance thresholds $d$ are based on the default values used to determine success in each environment. For fair comparison with methods that learn distances, including \abv, we relabel the distance between states $s_i$ and $s_j$ as 1, rather than the default value of $j - i$, if $r(s_i, a, \phi(s_j))$ = 0. Robomimic and Franka Kitchen, however, did not originally support goal-conditioning. Thus, we use the strict equality scheme as described below for images.

\textbf{Image.}
For image domains, we do not assume access to a ground-truth sparse reward function, because it is non-trivial to assess if a state is considered to have reached a goal given only image observations. Instead, we assign rewards to relabeled trajectories based only on transitions in the data, e.g. $r(s, a, \phi(s')) = 0$ if and only if $s'$ is the state that immediately follows $s$ in its trajectory, and $-1$ otherwise.

\subsection{Image Augmentation}
For our image-based experiments, we apply random shift augmentation as done in \citet{yarats2021improving}. For each training transition (state, next state, goal), we apply the same sampled augmentation to each image, and randomize augmentations across transitions.

\begin{table}[]
\resizebox{\textwidth}{!}{
\begin{tabular}{ccccccccccc}
\textbf{Environment}    & \textbf{Horizon} & \textbf{Learning Rate} & \textbf{Batch Size} & \textbf{Network Arch }       & \textbf{Train Steps} & \textbf{Eval Freq} & \textbf{Eval Ep} & \textbf{Plot Window} & \textbf{DWSL $N$} & \textbf{DWSL $B$} \\ \hline
Fetch State    & 50      & 0.0005        & 512        & {[}256, 256, 256{]} & 200k           & 2000      & 20      & 10          & 1           & 50        \\
Fetch Image    & 50      & 0.0003        & 256        & DrQv2               & 1m             & 5000      & 100     & 15          & 1           & 50        \\
Encoder ablation   & 50      & 0.0003        & 256        & ResNet18+{[}1024, 1024{]}               & 400k             & 5000      & 100     & 10          & 1           & 50        \\
Hand State     & 50      & 0.0005        & 512        & {[}256, 256, 256{]} & 200k           & 2000      & 20      & 10          & 1           & 50        \\
Hand Image     & 50      & 0.0003        & 256        & DrQv2               & 400k           & 5000      & 50      & 10          & 1           & 50        \\
Franka Kitchen & 280     & 0.0005        & 512        & {[}512, 512{]}      & 400k           & 10000     & 20      & 4           & 2           & 140       \\
Robomimic PH   & 500     & 0.0005        & 512        & {[}512, 512{]}      & 100k           & 5000      & 25      & 2           & 2           & 80        \\
Robomimic MH   & 500     & 0.0005        & 512        & {[}512, 512{]}      & 100k           & 5000      & 25      & 2           & 2           & 160       \\
Antmaze Umaze  & 700     & 0.0005        & 512        & {[}512, 512{]}      & 80k            & 1000      & 50      & 8           & 3           & 100       \\
Antmaze Medium & 900     & 0.0005        & 512        & {[}512, 512{]}      & 80k            & 1000      & 50      & 8           & 3           & 150       \\
Antmaze Large  & 1000    & 0.0005        & 512        & {[}512, 512{]}      & 80k            & 1000      & 50      & 8           & 3           & 200      
\end{tabular}
}
\caption{Domain Specific Hyperparameters} 
\label{tab:domain_hparams}
\end{table}

\end{document}